\documentclass[10pt,twocolumn,letterpaper]{article}
\usepackage[T1]{fontenc}
\usepackage[pagenumbers]{cvpr}
\usepackage{array}

\usepackage{amsmath,amssymb,booktabs,graphicx,cuted}
\usepackage[breaklinks,hidelinks]{hyperref}
\ifPDFTeX\pdfminorversion=7\fi
\ifXeTeX\AtBeginDvi{}\fi
\def\confName{CVPR}
\def\confYear{2027}
\hypersetup{
  pdftitle={Raysemble: Metric 3D Bone Reconstruction from a Single X-ray},
  pdfauthor={Xiaoyang Li; Xinghao Yan; Yuan Chai},
  pdfsubject={Research preprint},
  pdfkeywords={Single-view reconstruction, X-ray, bone reconstruction, projection geometry}
}
\title{Raysemble: Metric 3D Bone Reconstruction from a Single X-ray}
\author{
Xiaoyang Li$^{1}$ \qquad Xinghao Yan$^{1}$ \qquad Yuan Chai$^{1,2,*}$\\
{\normalsize $^{1}$ORBIT Lab, College of Medicine and Biomedical Information Engineering,}\\
{\normalsize Northeastern University, Shenyang 110016, China}\\
{\normalsize $^{2}$The John Curtin School of Medical Research, College of Science and Medicine,}\\
{\normalsize Australian National University, Canberra 2601, Australia}\\
{\small\href{mailto:20246389@stu.neu.edu.cn}{\texttt{20246389@stu.neu.edu.cn}}\quad
\href{mailto:yanxh@mails.neu.edu.cn}{\texttt{yanxh@mails.neu.edu.cn}}\quad
\href{mailto:yuanchai95@gmail.com}{\texttt{yuanchai95@gmail.com}}}\\
{\footnotesize ORCID:\quad
\href{https://orcid.org/0009-0002-4863-5761}{0009-0002-4863-5761}\quad
\href{https://orcid.org/0009-0002-4437-957X}{0009-0002-4437-957X}\quad
\href{https://orcid.org/0000-0001-6977-6155}{0000-0001-6977-6155}}\\
{\small $^{*}$Corresponding author: Yuan Chai}
}
\begin{document}
\raggedbottom
\maketitle
\begin{abstract}
Reconstructing bones from routine radiographs could support 3D orthopedic assessment. We present Raysemble, which separates shape and placement to recover eight lumbar and pelvic bones from a single X-ray. Each 3D query samples local and full-frame features at the same projected point. A shared decoder predicts normalized bone shapes. A separate branch estimates depth, back-projects centers, and adjusts scale for magnification. Resampling assembles the bones in physical coordinates. On 43 simulated anteroposterior radiographs from 23 held-out patients, Raysemble ranks first among 15 configurations across three seeds before alignment, achieving 65.24\% Dice, 9.09\,mm HD95, and 3.32\,mm ASSD. With identical inputs, it improves Dice over VolumeNeRF by 5.86 percentage points.
\end{abstract}

\begin{figure*}[t]
  \centering
  \includegraphics[width=\textwidth]{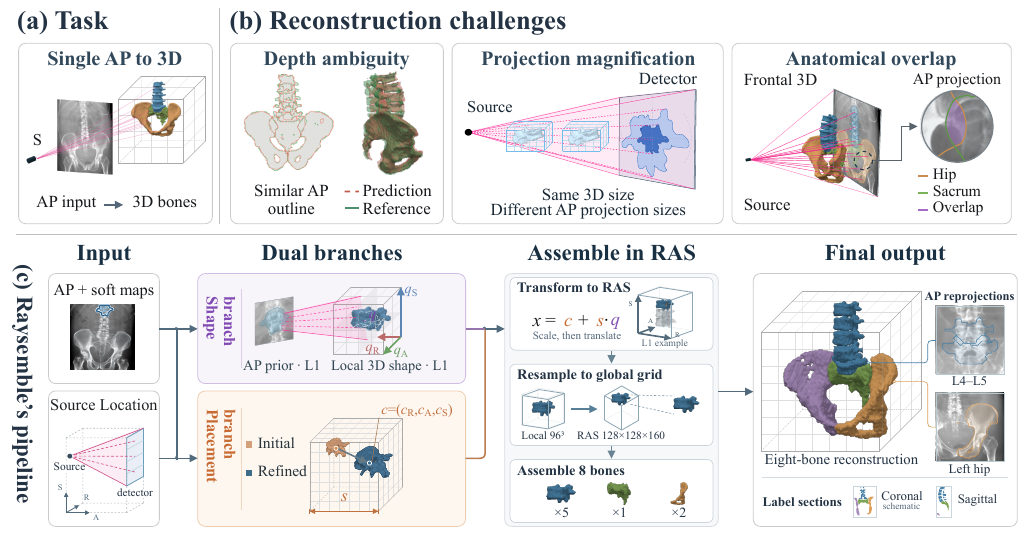}
  \caption{From one AP radiograph to eight bones. \textbf{(a)} The reconstruction task. \textbf{(b)} Depth ambiguity, projection magnification, and anatomical overlap. \textbf{(c)} Raysemble estimates normalized bone shape and metric placement in two branches. The predicted center and scale transform local occupancies into the shared right--anterior--superior (RAS) frame through \(\mathbf x=\mathbf c+s\mathbf q\). Resampling and fusion assemble the eight-bone reconstruction.}
  \label{fig:teaser}
\end{figure*}

\section{Introduction}
\label{sec:intro}
From a single anteroposterior (AP) radiograph, we aim to recover the surfaces, physical sizes, and relative positions of L1--L5, the sacrum, and both hip bones (Fig.~\ref{fig:teaser}a). Similar projected outlines can conceal different depths, cone-beam magnification couples depth and apparent size, and lumbar and pelvic structures overlap in the image (Fig.~\ref{fig:teaser}b). Recovering this anatomy requires both local bone detail and the spatial relationships among bones.

Methods that reconstruct a 3D volume from X-rays predict anatomy on a common grid~\cite{ying2019,volumenerf2024,bai2024}. Statistical shape models constrain the range of plausible anatomy~\cite{zheng2010,picard2026}. A normalized bone shape can occur at different positions and physical sizes. Raysemble uses this distinction to estimate shape and placement separately, with camera geometry relating depth to magnification.

The two branches meet during assembly (Fig.~\ref{fig:teaser}c). An initial placement positions the 3D queries used to reconstruct each bone. Each query is projected once to sample features from both the bone crop and the full image, preserving local detail and anatomical context. A shared decoder predicts occupancy on a normalized grid. The final placement branch estimates depth, back-projects bone centers, and adjusts scale for magnification. Resampling combines the decoded shape with the predicted center and scale on a common grid in millimeters (Fig.~\ref{fig:overview}).

We compare 15 configurations before alignment and test Raysemble against VolumeNeRF with identical images, probability maps, bone regions, and camera parameters. Raysemble leads on Dice, HD95, and ASSD. A fixed-shape comparison isolates the mean improvement from final placement, while separate evaluations examine rigid alignment and exploratory real pelvic radiographs.

\section{Related Work}
\paragraph{CT reconstruction from X-rays.}
X2CT-GAN synthesizes CT from two orthogonal radiographs, and SAX-NeRF reconstructs radiodensity from sparse projections~\cite{ying2019,saxnerf2024}. With a single radiograph, Henzler et al.\ first predict a coarse CT volume and then use the input image to refine it~\cite{henzler2018}. VolumeNeRF conditions CT synthesis on an anatomical likelihood, while XctDiff uses priors extracted from the radiograph to guide latent diffusion~\cite{volumenerf2024,bai2024}. These methods target an entire volume of CT intensities. To obtain bone labels, we pair the synthesized volume with a segmenter. Our matched-input comparison tests this route against predicting bone occupancy directly.

\paragraph{Bone reconstruction from one or two views.}
Zheng uses landmark-to-ray correspondences to initialize pelvic pose and scale, then iteratively fits a statistical shape model to the projected contours~\cite{zheng2010}. The study estimates unknown model scales by matching the reconstructed surface to its CT reference. Joint 2D--3D SSM learns a shared PCA space for deformations of 2D masks and 3D shapes, allowing it to infer the 3D deformation from a 2D observation~\cite{picard2026}. 3DDX instead predicts depths for the front and back surfaces, uses the camera geometry to reconstruct an initial surface, and fits a statistical shape model to complete it~\cite{gu2024}. X2V and X2BR represent organ and bone geometry, respectively, with implicit fields~\cite{guven2024,x2br2025}.

A second view provides complementary anatomical evidence. Lin et al.\ reconstruct multiple leg bones from biplanar X-rays, and X2Bone fuses AP and lateral views to reconstruct the pelvis~\cite{lin2026,pan2026x2bone}. For a single AP view, Raysemble must infer each bone's depth from the image and learned anatomical relationships. It predicts bone shape on a normalized grid and estimates the center and scale separately for assembly.

\paragraph{Projection-aligned features.}
PIFu samples image features at projected 3D points~\cite{pifu2019}. DISN combines local samples with a global image descriptor, and PIFuHD uses features from a coarser stage to guide fine prediction~\cite{disn2019,pifuhd2020}. For sparse-view CBCT, C$^2$RV combines features across scales and views~\cite{lin2024c2rv}. Raysemble projects a query into a single AP image, then samples two spatial feature maps at that point. The crop provides detail around the target bone, and the full-image map preserves the surrounding anatomy.

\paragraph{Normalized shape and physical placement.}
Object reconstruction also uses normalized coordinates to separate shape from placement. NOCS estimates pixel correspondences to a shared object coordinate system and uses measured depth to recover metric pose and size~\cite{wang2019nocs}. In our setting, the image center specifies a ray through the X-ray source. Raysemble estimates depth along that ray to place the bone center, then uses the change in magnification to adjust its scale. This separation lets us test placement changes with the decoded bone shapes held fixed.

\section{Method}
\label{sec:method}
\begin{figure*}[t]
  \centering
  \includegraphics[width=\textwidth]{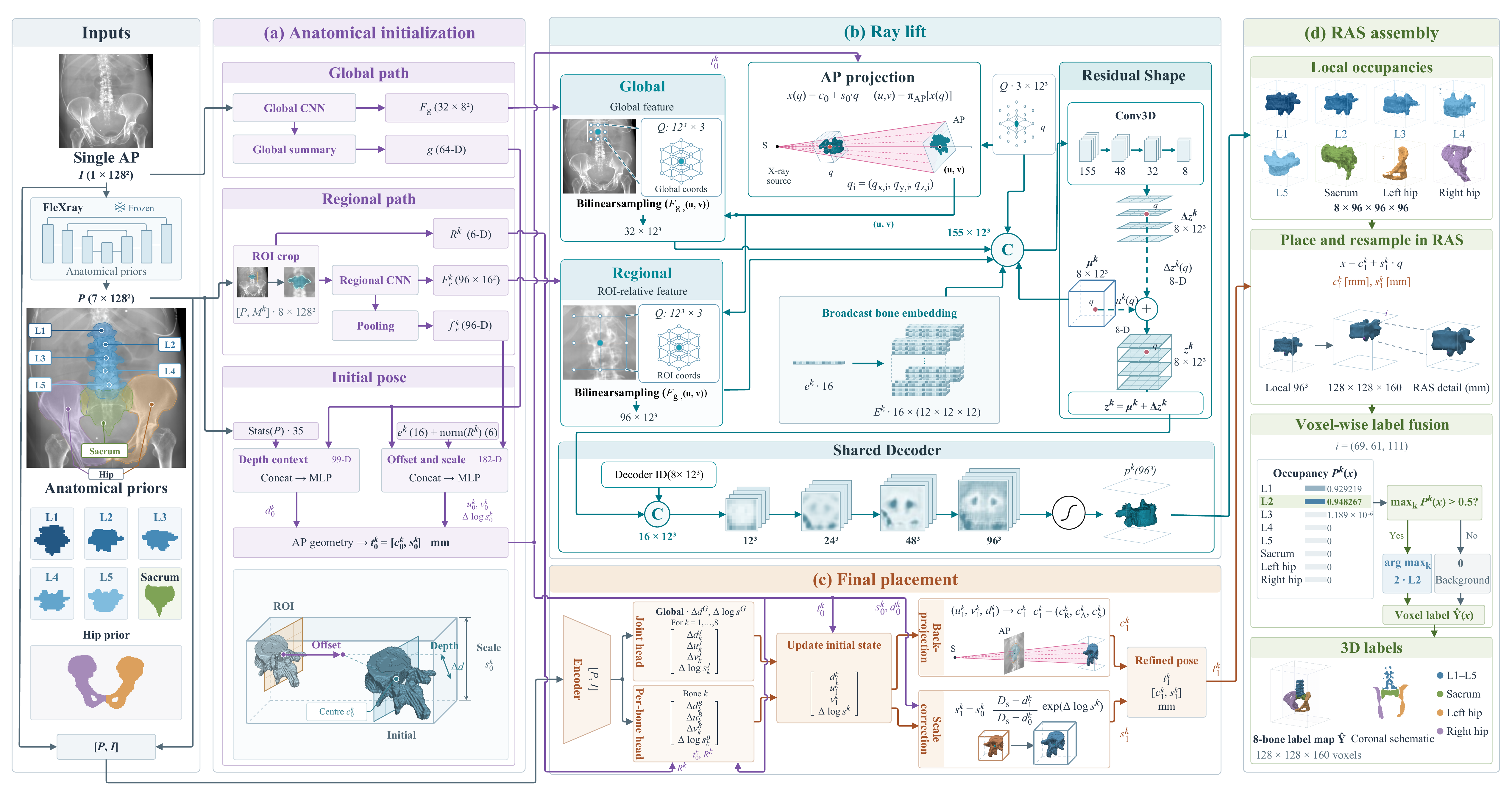}
  \caption{Raysemble architecture. \textbf{(a)} Global and regional probability-map features initialize each bone's query center and scale. \textbf{(b)} Each query is projected once: the same detector coordinate \((u,v)\) samples the global and ROI-relative feature maps. A residual shape predictor and shared decoder recover local occupancy through \(12^3\), \(24^3\), \(48^3\), and \(96^3\) grids. \textbf{(c)} The placement branch predicts depth, back-projects centers, and corrects scale for magnification. \textbf{(d)} Local occupancies and final centers and scales supply separate assembly inputs. Resampling onto the common RAS grid and voxel-wise fusion produce eight bone labels.}
  \label{fig:overview}
\end{figure*}
Raysemble uses two estimates of each bone's placement. The initial center \(\mathbf c^0_k\) and side length \(s^0_k\) position the queries for shape reconstruction (Fig.~\ref{fig:overview}a,b). The final center \(\mathbf c^1_k\) and side length \(s^1_k\) place the decoded occupancy in physical space (Fig.~\ref{fig:overview}c,d).

\subsection{Anatomical initialization}
\paragraph{Projection geometry.}
We use a right--anterior--superior (RAS) coordinate system centered at the isocenter. For a point \(\mathbf x=(x,d,z)\), depth \(d\) increases toward the source. Given source-to-isocenter distance \(D_s\), source-to-detector distance \(D_d\), detector pixel size \(\rho\), and detector center \((u_0,v_0)\), its projection is
\begin{equation}
 \pi(\mathbf{x})=
 \left(u_0-\frac{D_dx}{\rho(D_s-d)},\ 
 v_0-\frac{D_dz}{\rho(D_s-d)}\right).
 \label{eq:projection}
\end{equation}
The detector spans \(450\,\mathrm{mm}\) per side, with \(D_s=1050\,\mathrm{mm}\) and \(D_d=1200\,\mathrm{mm}\). We estimate the torso isocenter from CT intensities, map it to RAS using the CT affine, and center a fixed 3D crop there to render the AP projection. The camera and crop are determined independently of target bone annotations (Supplementary Sec.~\ref{sec:supp_coordinates}).

\paragraph{Bone probability maps.}
The frozen FleXray segmenter~\cite{flexray2026} processes the \(512\times512\) AP image and separate lumbar and pelvic crops without test-time augmentation. We restore the probability maps to detector coordinates and area-downsample them to \(P\in\mathbb R^{7\times128\times128}\), covering L1--L5, the sacrum, and a combined hip class. Splitting the hip map at the predicted sacral centroid gives eight target maps \(M_k\). For each bone, we enclose the largest connected region above 0.25 probability in a square ROI, enlarge its side by 1.35, and enforce a 12-pixel minimum. A missing region triggers a full-image crop. The encoders retain the soft probabilities. \(R_k\) records the ROI center, side length, width, height, and probability mass.

\paragraph{Initialize the query grid.}
As shown in Fig.~\ref{fig:overview}a, the full-frame encoder maps \(P\) to \(F^g\in\mathbb R^{32\times8\times8}\) and a 64-dimensional descriptor \(g\). The regional encoder maps a crop of \([P,M_k]\) to \(F^r_k\in\mathbb R^{96\times16\times16}\). Pooled regional features, \(R_k\), a bone embedding, and \(g\) determine the initial image offset and scale. A depth head takes \(g\) and five probability moments per class to predict \(d^0_k=\bar d_k+30\tanh(a^0)+20\tanh(b^0_k)\), where \(\bar d_k\) is a training-set template. The shared and per-bone corrections are bounded by 30 and 20\,mm, respectively. Back-projection gives \((\mathbf c^0_k,s^0_k)\).

\subsection{Ray lifting and bone shape}
\paragraph{One projection, two feature samples.}
For a normalized query \(\mathbf q\in[-0.5,0.5]^3\) on the \(12^3\) grid, the detector location is
\begin{equation}
 (u,v)=\pi\!\left(\mathbf c^0_k+s^0_k\mathbf q\right).
 \label{eq:query}
\end{equation}
Figure~\ref{fig:overview}b shows the global and regional samplers above and below, respectively. We express the same \((u,v)\) in each map's coordinates and sample by bilinear interpolation. The 32 global channels preserve surrounding anatomy, while the 96 regional channels resolve local detail. We concatenate both samples with a 16-dimensional bone embedding, the three coordinates of \(\mathbf q\), and eight channels from the mean latent \(\mu_k\). A \(155\!\rightarrow\!48\!\rightarrow\!32\!\rightarrow\!8\) 3D convolutional head predicts \(\Delta z_k\), giving \(z_k=\mu_k+\Delta z_k\).

\paragraph{Decode normalized occupancy.}
The shared decoder concatenates \(z_k\) with an eight-channel bone embedding and upsamples through \(24^3\), \(48^3\), and \(96^3\) grids. A sigmoid produces local occupancy probabilities \(p_k\). Training targets retain their RAS orientation within a cube centered on each bone's bounding box. Its side is \(1.12\) times the longest box dimension plus twice the largest original voxel spacing. We normalize the cube to \([-0.5,0.5]^3\). Autoencoder training, mean latents, and depth templates use training patients only.

\subsection{Final placement and RAS assembly}
\begin{table*}[t]
\centering
\caption{Eight-bone reconstruction before alignment on 23 test patients. Means are patient-weighted and the standard deviation is across three seeds. Total inference parameters include the frozen FleXray front end where used. Bold marks the best mean in each metric. Per-bone results are in Supplementary Table~\ref{tab:per_bone}.}
\label{tab:main}
\small
\setlength{\tabcolsep}{9pt}
\begin{tabular*}{\textwidth}{@{\extracolsep{\fill}}lrrrr@{}}
\toprule
Method & Params. (M) & Dice (\%) $\uparrow$ & HD95 (mm) $\downarrow$ & ASSD (mm) $\downarrow$ \\
\midrule
Broadcast U-Net & 36.35 & $56.45 \pm 1.17$ & $14.19 \pm 2.40$ & $5.01 \pm 0.63$ \\
Henzler & 8.08 & $52.46 \pm 0.53$ & $19.85 \pm 4.19$ & $9.28 \pm 4.27$ \\
PatRecon & 592.52 & $42.39 \pm 6.25$ & $27.33 \pm 8.77$ & $12.50 \pm 5.44$ \\
XctNet & 33.94 & $52.60 \pm 2.16$ & $25.35 \pm 12.62$ & $14.35 \pm 12.46$ \\
3DDX-Direct & 142.54 & $56.08 \pm 0.95$ & $14.35 \pm 0.95$ & $4.73 \pm 0.11$ \\
X2V & 153.39 & $41.21 \pm 1.23$ & $25.03 \pm 0.80$ & $10.15 \pm 0.20$ \\
X2B & 95.80 & $40.35 \pm 1.19$ & $27.58 \pm 4.39$ & $12.39 \pm 4.03$ \\
SR-Net & 22.09 & $54.32 \pm 1.30$ & $15.58 \pm 1.28$ & $5.38 \pm 0.27$ \\
XctDiff & 62.16 & $31.22 \pm 8.82$ & $36.28 \pm 16.74$ & $17.03 \pm 10.62$ \\
VolumeNeRF & 109.42 & $45.72 \pm 5.65$ & $23.36 \pm 2.13$ & $8.28 \pm 1.02$ \\
RT-SRTS & 624.02 & $30.23 \pm 14.20$ & $137.56 \pm 63.21$ & $32.21 \pm 16.28$ \\
Joint 2D--3D SSM & 2.25 & $35.99 \pm 1.99$ & $28.40 \pm 4.13$ & $7.75 \pm 0.80$ \\
SSR/PIFu & 116.07 & $54.48 \pm 0.21$ & $16.84 \pm 0.21$ & $5.84 \pm 0.03$ \\
VolumeNeRF (matched) & 210.83 & $59.38 \pm 0.30$ & $19.63 \pm 6.49$ & $5.76 \pm 1.25$ \\
\textbf{Raysemble} & 102.19 & $\mathbf{65.24 \pm 0.46}$ & $\mathbf{9.09 \pm 0.15}$ & $\mathbf{3.32 \pm 0.06}$ \\
\bottomrule

\end{tabular*}
\end{table*}

\paragraph{Predict depth and center.}
The placement encoder in Fig.~\ref{fig:overview}c takes \([P,I]\), where \(I\) is the AP image resized to \(128\times128\). A joint head predicts shared and per-bone adjustments. The local head also receives bone identity, \(R_k\), and the initial placement with depth masked. With training-set depth template \(\bar d_k\), shared output \(a\), and combined per-bone output \(b_k\), absolute depth is
\begin{equation}
 d^1_k=\bar d_k+80\tanh(a)+40\tanh(b_k),
 \label{eq:depth}
\end{equation}
where bounds are in millimeters. The shared term shifts the skeleton along the AP axis, and the per-bone term adjusts relative depth. We update the initial image center with predicted offsets and back-project at \(d^1_k\) to obtain \(\mathbf c^1_k\).

\paragraph{Correct scale for magnification.}
For a fixed projected outline, a bone farther from the source must be physically larger (Fig.~\ref{fig:teaser}b). Equation~\eqref{eq:projection} gives the scale update
\begin{equation}
 s^1_k=s^0_k\,
 \frac{D_s-d^1_k}{D_s-d^0_k}\,\exp(\delta_k),
 \label{eq:scale}
\end{equation}
with a bounded log-scale residual \(\delta_k\). At a fixed image center and \(\delta_k=0\), this rescales the bone about the source while preserving its projection before side-length clipping. Anatomical cues determine depth, and camera geometry links that depth to scale. The physical update uses \(d^0_k\), although the local head's input masks it.

\paragraph{Assemble the bones.}
In Fig.~\ref{fig:overview}d, \(p_k\) supplies the local shape and \((\mathbf c^1_k,s^1_k)\) supplies its placement to Resample. We transform local points by \(\mathbf x=\mathbf c^1_k+s^1_k\mathbf q\), preserve RAS orientation, and trilinearly resample occupancy onto a \(128\times128\times160\) grid. Voxel centers span \((-239,-143,-287)\) to \((239,143,287)\)\,mm at approximately \((3.764,2.252,3.610)\)\,mm spacing. Each voxel takes the most probable bone label above 0.5, otherwise background. The grid is fixed from training data. Inference uses the AP image and camera geometry.

\subsection{Training}
Each seed has three stages of 60 epochs, each covering eligible samples once. We first train the autoencoder on normalized CT masks. With its decoder frozen, we train the image encoders, shape predictor, and initial placement. Supervision covers local occupancy, latent codes, mask overlap in physical space, projected silhouettes, and relative depth. We then freeze the shape predictor and train final placement using center, log side length, mask overlap, and silhouettes, emphasizing bones with larger initial center errors. Absolute depth prediction is introduced over the first five epochs.

\section{Experiments}
\label{sec:experiments}
\subsection{Data and evaluation}
\paragraph{Patient splits and AP images.}
We use 419 CT scans from 225 subjects in the CT-TRI and AbdomenCT-1K subsets of CADS~\cite{cads2026,abdomen2021}. We generate one digitally reconstructed AP radiograph (DRR) per scan with DiffDRR~\cite{gopalakrishnan2022fast} using fixed cone-beam geometry. The training, validation, and test splits contain 338, 38, and 43 scans from 180, 22, and 23 subjects, respectively. We keep all scans from a subject in the same split. The test set contains 34 CT-TRI and nine AbdomenCT-1K scans. We use TotalSegmentator~\cite{totalsegmentator2023} to generate the reference masks and map them to the eight target bone labels. We screen scans for label presence, bone volume, field of view, and metal artifacts (Supplementary Sec.~\ref{sec:supp_coordinates}).

\paragraph{DRR generation.}
We use the Siddon renderer with zero yaw, pitch, and roll. The source-to-isocenter and source-to-detector distances are 1050 and 1200\,mm, respectively, with a $450\times450$\,mm detector. We render at $1024\times1024$ pixels, normalize each image between its 1st and 99th intensity percentiles, and bilinearly resize it to $512\times512$ pixels. Raysemble uses $128\times128$ AP images for 3D reconstruction; its FleXray front end processes the $512\times512$ images. Supplementary Sec.~\ref{sec:supp_coordinates} gives the density conversion and coordinate mapping.

\paragraph{Metrics and alignment.}
We report Dice, the symmetric 95th percentile Hausdorff distance (HD95), and average symmetric surface distance (ASSD). For the main comparison, we score the predictions before alignment to CT and measure distances in millimeters on the fixed RAS grid. We first average over the eight bones, then over each patient's scans, and finally across patients with equal weights. In a separate comparison, we align the entire reconstruction to CT using rotation and translation only. If a reference bone is missing from the prediction, we assign a Dice score of zero and use the grid diagonal for both distance metrics.

\paragraph{Single-view baselines.}
The direct baseline broadcasts 2D features into a 3D U-Net. Projection-to-volume adaptations include Henzler~\cite{henzler2018}, PatRecon~\cite{shen2019}, XctNet~\cite{tan2022}, XctDiff~\cite{bai2024}, and VolumeNeRF~\cite{volumenerf2024}. Bone- and shape-centered adaptations include 3DDX-Direct~\cite{gu2024}, X2V~\cite{guven2024}, X2B~\cite{x2br2025}, SR-Net~\cite{huang2026}, RT-SRTS~\cite{zhu2024}, Joint 2D--3D SSM~\cite{picard2026}, and SSR/PIFu~\cite{pifu2019,lin2026}. We adapt each method to use one AP view and predict eight bone labels. Each trainable stage runs for 60 epochs. Supplementary Table~\ref{tab:baseline_protocol} lists inputs, supervision, adaptations, training stages, and checkpoint selection for all 15 configurations.

\subsection{Reconstruction before alignment}
Before alignment, Raysemble achieves the highest Dice and lowest surface distances, with 65.24\% Dice, 9.09\,mm HD95, and 3.32\,mm ASSD (Table~\ref{tab:main}). Relative to the best comparator on each metric, it gains 5.86 Dice points over matched VolumeNeRF, lowers HD95 by 5.10\,mm over Broadcast U-Net, and lowers ASSD by 1.40\,mm over 3DDX-Direct.

\begin{figure*}[p]
 \centering
 \newsavebox{\syntheticArtwork}
 \sbox{\syntheticArtwork}{\includegraphics[width=\textwidth,height=.93\textheight,keepaspectratio]{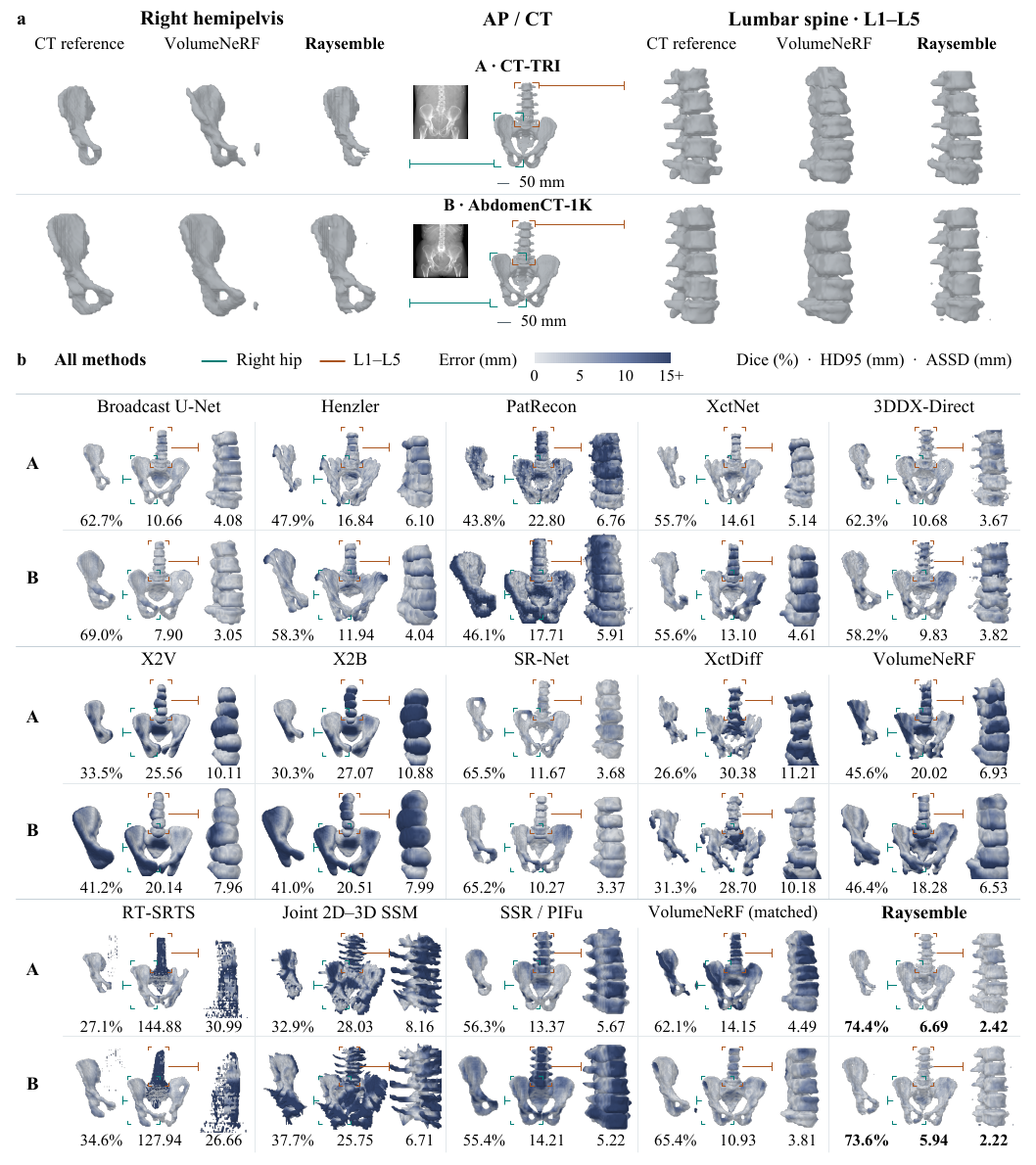}}
 \usebox{\syntheticArtwork}
 \caption{Synthetic AP reconstruction. \textbf{(a)} CT, input-matched VolumeNeRF, and Raysemble. \textbf{(b)} Anterior views of the whole anatomy and right hip, with 25-degree anterior oblique views of L1--L5, for 15 configurations. Colors show distances from seed-0 predictions to the nearest CT vertex of the corresponding bone (0--15\,mm). Bold marks the best three-seed means per case.}
 \label{fig:synthetic}
 \vspace{\dimexpr.93\textheight-\ht\syntheticArtwork\relax}
\end{figure*}

Figure~\ref{fig:synthetic} compares one case from each CT source. Panel (a) groups the right hip and L1--L5 at a common scale. Panel (b) shows all methods with anterior views of the full anatomy and right hip, alongside 25-degree anterior oblique views of L1--L5. CT-centered windows make local shape and placement directly comparable.

\begin{figure*}[t]
 \centering
 \includegraphics[width=\textwidth]{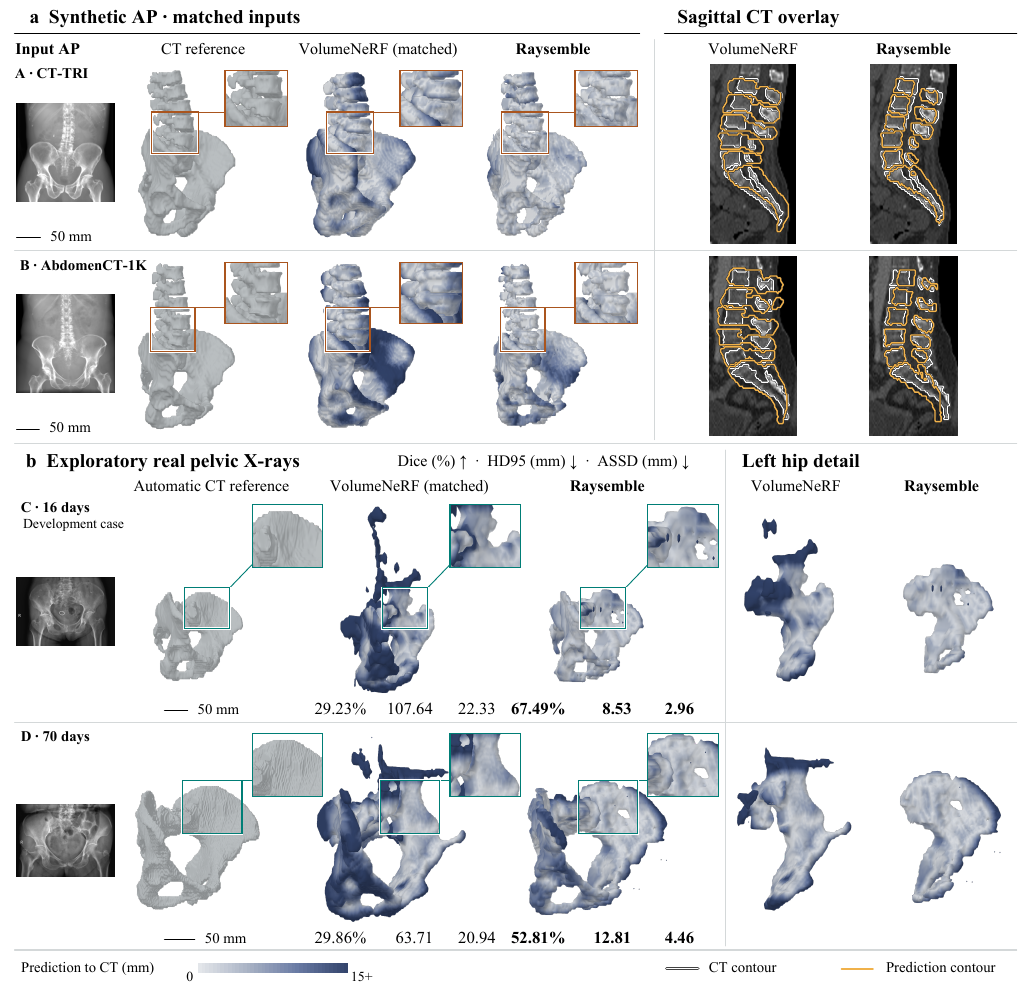}
 \caption{Reconstruction with identical inputs. \textbf{(a)} Synthetic AP cases before alignment. \textbf{(b)} Real pelvic X-rays with X-ray--CT intervals of 16 and 70 days, rigidly aligned to CT reference masks without scaling. Insets share CT-defined L4--L5 or iliac-wing windows. Colors show distances from seed-0 predictions to the nearest CT vertex of the corresponding bone (0--15\,mm). Bold marks the best three-seed mean for each real-case metric.}
 \label{fig:matched_visual}
\end{figure*}

\subsection{Reconstruction with matched inputs}
We give VolumeNeRF and Raysemble the same AP image, FleXray probability maps, bone regions, and camera parameters. The adapted VolumeNeRF retains the original generator and deformable attention~\cite{volumenerf2024} as well as the anatomical likelihood estimated from training data. We use a fixed segmenter, trained on the training split, to label the synthesized CT.

\begin{table}[t]
\centering
\caption{Paired differences from VolumeNeRF with matched inputs on 23 patients. Positive values favor Raysemble. We compute confidence intervals from 10,000 patient-level bootstrap resamples and round only after computing the differences.}
\label{tab:matched_pairs}
\footnotesize
\setlength{\tabcolsep}{3pt}
\begin{tabular}{@{}lrrr@{}}
\toprule
Metric & Mean gain & 95\% CI & Improved \\
\midrule
Dice (pp) & +5.86 & [3.35, 8.94] & 18/23 \\
HD95 (mm) & +10.55 & [5.34, 18.82] & 23/23 \\
ASSD (mm) & +2.44 & [1.37, 4.36] & 23/23 \\
\bottomrule

\end{tabular}
\end{table}

We compare each patient's scores after averaging over scans and the three seeds. Raysemble improves Dice for 18 of 23 patients and lowers both distance metrics for all 23 (Table~\ref{tab:matched_pairs}). HD95 and ASSD fall by 10.55 and 2.44\,mm on average, with median reductions of 4.98 and 1.53\,mm. The mean reductions are 7.14 and 1.56\,mm after excluding the patient with the largest gain. Both methods predict all eight bones for every test scan (Supplementary Sec.~\ref{sec:supp_failures}).

Figure~\ref{fig:matched_visual}a separates local shape and placement visually through shared L4--L5 windows and sagittal overlays against CT.

\paragraph{Real pelvic radiographs.}
We also examine two real pelvic AP radiographs from a single hospital, with X-ray--CT examination intervals of 16 and 70 days. The models use crops from the radiographs and nominal camera settings. After rigid alignment without scaling to the automatically segmented CT pelvis, Raysemble leads on all three metrics in both cases (Fig.~\ref{fig:matched_visual}b). The clinical analysis received ethics approval and a waiver of informed consent (Supplementary Sec.~\ref{sec:supp_ethics}).

\paragraph{Per-bone accuracy and parameters.}
Raysemble has the lowest HD95 and ASSD on all eight bones and the highest Dice on seven (Supplementary Table~\ref{tab:per_bone}). Broadcast U-Net has slightly higher sacrum Dice (62.36\% versus 61.94\%), while Raysemble reduces sacrum HD95 from 13.81 to 10.88\,mm and ASSD from 4.90 to 4.18\,mm. Raysemble has 0.79 million parameters in its reconstruction network. With the shared FleXray model included, Raysemble and the matched-input VolumeNeRF have 102.19 and 210.83 million parameters, respectively.

\subsection{Shape and placement ablations}
\begin{table}[t]
\centering
\caption{Ablations on the test set before alignment. Scores are means $\pm$ standard deviations across seeds. The first three rows share the shape predictor. Validation selected the residual model at epoch 0 in all seeds. The overlap loss compares masks after placement.}
\label{tab:ablation}
\footnotesize
\setlength{\tabcolsep}{2pt}
\begin{tabular}{@{}lrrr@{}}
\toprule
Method & Dice (\%) $\uparrow$ & HD95 (mm) $\downarrow$ & ASSD (mm) $\downarrow$ \\
\midrule
Initial placement & $63.55 \pm 0.72$ & $9.63 \pm 0.26$ & $3.50 \pm 0.10$ \\
Residual placement & $63.55 \pm 0.72$ & $9.63 \pm 0.26$ & $3.50 \pm 0.10$ \\
\textbf{Absolute placement} & $\mathbf{65.24 \pm 0.46}$ & $\mathbf{9.09 \pm 0.15}$ & $\mathbf{3.32 \pm 0.06}$ \\
\midrule
\multicolumn{4}{@{}l}{\textit{Shape stage}} \\
Raw, full frame & $53.73 \pm 0.41$ & $13.70 \pm 0.25$ & $5.58 \pm 0.11$ \\
Raw, predicted ROI & $62.94 \pm 0.51$ & $10.32 \pm 0.10$ & $3.69 \pm 0.05$ \\
Bone probability maps & $63.55 \pm 0.72$ & $9.63 \pm 0.26$ & $3.50 \pm 0.10$ \\
No overlap loss & $61.49 \pm 0.89$ & $10.20 \pm 0.32$ & $3.72 \pm 0.12$ \\
No shape pretraining & $65.28 \pm 1.28$ & $9.46 \pm 0.21$ & $3.38 \pm 0.09$ \\
\bottomrule
\end{tabular}
\end{table}

\paragraph{Placement with shape held fixed.}
All three variants use the same shape predictor. Compared with the initial placement, absolute depth prediction improves mean Dice before alignment by 1.69 points (paired 95\% CI $-0.60$ to 3.92) and reduces HD95 by 0.54\,mm and ASSD by 0.18\,mm (Tab.~\ref{tab:ablation}). Validation selects the residual checkpoint at epoch 0 for all three seeds. Its head is still zero-initialized, so it returns nearly the same placement as the initial estimate. The designs differ in depth range, use of the initial depth, and training augmentation.

AP center error falls by 0.59\,mm, with little change in log side-length error (Supplementary Fig.~S1). Paired confidence intervals for both errors and surface distances include zero. After rigid alignment, the Dice gain is 0.82 points.

\paragraph{Shape inputs and supervision.}
Cropping the raw image to the predicted ROI improves all three metrics, and replacing those crops with bone probability maps further reduces surface errors. Removing the overlap loss after placement worsens all three metrics. The variant without shape pretraining uses the initial placement and reaches 65.28\% Dice before alignment, compared with 65.24\% for the final model. It also improves on the base shape model. This variant changes decoder initialization, the mean latent code, and latent supervision together.

\subsection{Accuracy across CT sources and alignment}
\label{sec:robust}
\begin{table}[t]
\centering
\caption{Reconstruction with identical inputs, by CT source and alignment. Each cell gives VolumeNeRF / Raysemble. Means weight patients equally across three seeds.}
\label{tab:matched_strata}
\footnotesize
\setlength{\tabcolsep}{2.5pt}
\begin{tabular}{@{}lrrr@{}}
\toprule
Test group & Dice (\%) & HD95 (mm) & ASSD (mm) \\
\midrule
\multicolumn{4}{@{}l}{\textit{Before alignment}} \\
CT-TRI (14) & 57.39 / \textbf{63.68} & 24.03 / \textbf{9.53} & 6.63 / \textbf{3.50} \\
AbdomenCT-1K (9) & 62.49 / \textbf{67.66} & 12.80 / \textbf{8.40} & 4.42 / \textbf{3.05} \\
\midrule
\multicolumn{4}{@{}l}{\textit{Rigid alignment without scaling}} \\
CT-TRI (14) & 60.63 / \textbf{72.85} & 19.87 / \textbf{7.02} & 5.73 / \textbf{2.54} \\
AbdomenCT-1K (9) & 66.39 / \textbf{74.71} & 11.12 / \textbf{6.85} & 3.92 / \textbf{2.42} \\
All patients (23) & 62.88 / \textbf{73.58} & 16.44 / \textbf{6.96} & 5.02 / \textbf{2.49} \\
\bottomrule
\end{tabular}
\end{table}

\paragraph{Accuracy across CT sources.}
With the same simulated AP camera, Raysemble improves mean Dice, HD95, and ASSD on both CT-TRI and AbdomenCT-1K (Table~\ref{tab:matched_strata}). Before alignment, Dice gains are 6.30 and 5.17 points, respectively. HD95 reductions vary more between sources. The means are 14.49 and 4.41\,mm, and the medians are 5.89 and 3.72\,mm. HD95 and ASSD are lower for all 14 CT-TRI patients and all nine AbdomenCT-1K patients.

\paragraph{Accuracy after rigid alignment.}
After alignment, Raysemble improves all three metrics over the matched VolumeNeRF for every patient (Table~\ref{tab:matched_strata}). A single rigid transform per case corrects global position and orientation, leaving local shape, size, and relative bone positions to affect the remaining errors.

\paragraph{Lumbar and pelvic surfaces.}
Before alignment, Raysemble reduces mean ASSD relative to the matched VolumeNeRF by 2.67\,mm across L1--L5 and 2.06\,mm across the sacrum and both hips. The reductions after alignment are 2.48 and 2.61\,mm, respectively. The CT-defined insets locate these surface differences within shared anatomical windows.

\section{Discussion}
\paragraph{Bone shape and placement.}
The two branches address different ambiguities in Fig.~\ref{fig:teaser}b. Sampling local and full-frame features connects each bone to the surrounding anatomy. Separating normalized shape from placement makes it possible to update center and scale without changing the decoded surface. The advantage over matched VolumeNeRF persists after rigid alignment, which preserves differences in local shape, physical size, and relative bone positions.

\paragraph{What the comparisons establish.}
The matched-input comparison tests direct bone reconstruction against CT synthesis followed by segmentation. The same segmenter reaches 91.82\% Dice on reference CT and 59.38\% on synthesized CT (Supplementary Sec.~\ref{sec:supp_failures}). With shape held fixed, final placement improves mean reconstruction scores, although confidence intervals include zero and scale error changes little.

\paragraph{Acquisition and field of view.}
Clinical evaluation before alignment requires calibrated camera geometry and confirmed image pairs. Magnification errors can be mistaken for bone-size errors. FleXray's public training sources omit CT-TRI and AbdomenCT-1K, but patient-level independence requires identifiers absent from those records (Supplementary Sec.~\ref{sec:supp_coordinates}). Truncated bones can displace the predicted ROI and the initial query grid, affecting both feature samples. Assembly preserves normalized bone orientation. Clinical measurements require independently annotated references.

\section{Conclusion}
Raysemble reconstructs eight bones from a single AP radiograph by separating normalized shape from metric placement. Shared projected queries sample local and full-frame features, while magnification links center and scale. Raysemble leads the 15-configuration comparison before alignment and retains its matched-input advantage after rigid alignment.

\section*{Acknowledgments}
We thank Dr.\ Quan Zhou for coordinating the ethics approval process and the collection of the clinical X-ray and CT data at the First People's Hospital of Dali City.

\section*{LLM Usage Statement}
GPT-6.1 Sol (OpenAI Codex) assisted with manuscript editing and language polishing, visualization preparation, and code organization.
{\small
\bibliographystyle{ieeenat_fullname}
\bibliography{references}
}

\clearpage
\setcounter{section}{0}
\setcounter{figure}{0}
\setcounter{table}{0}
\setcounter{equation}{0}
\renewcommand{\thesection}{S\arabic{section}}
\renewcommand{\thefigure}{S\arabic{figure}}
\renewcommand{\thetable}{S\arabic{table}}
\renewcommand{\theequation}{S\arabic{equation}}
\renewcommand{\theHsection}{supp.\arabic{section}}
\renewcommand{\theHfigure}{supp.\arabic{figure}}
\renewcommand{\theHtable}{supp.\arabic{table}}
\renewcommand{\theHequation}{supp.\arabic{equation}}
\pdfbookmark[0]{Supplementary Material}{supplementary}
\captionsetup{skip=3pt}
\twocolumn[{
\begin{center}
{\Large\bfseries Supplementary Material}\par\vspace{5pt}
\includegraphics[width=\textwidth]{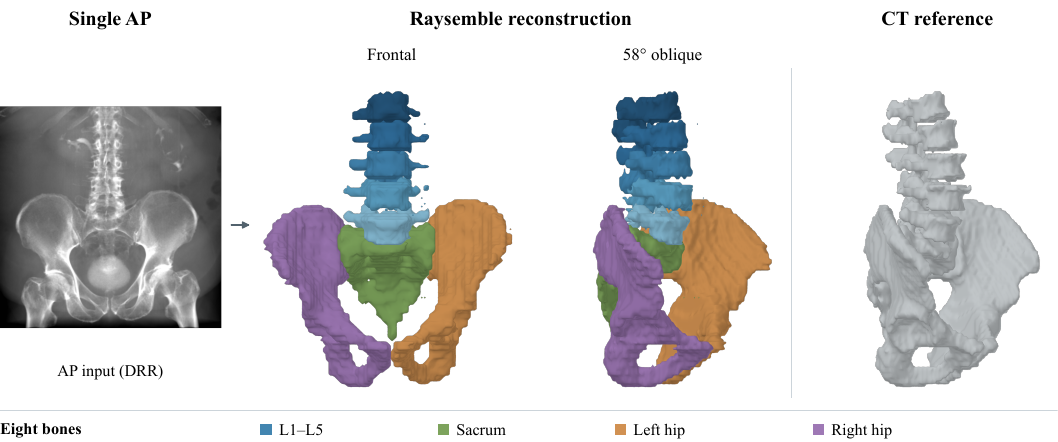}\par
\vspace{3pt}\small One CT-TRI case: AP input, eight-bone native reconstruction, and the matching CT reference.
\end{center}\vspace{4pt}
}]
\section{Coordinate and data protocol}
\label{sec:supp_coordinates}
\paragraph{AP acquisition.}
DiffDRR generates one AP per CT with Siddon rendering, zero rotation, source $(0,1050,0)$\,mm, and a $450\times450$\,mm detector at $y=-150$\,mm. Source--isocenter/detector distances are 1050/1200\,mm. We use 1\,mm CT resampling, HU-to-density conversion, bone attenuation multiplier 6, \texttt{center\_volume=False}, and \texttt{patch\_size=128}. Rendering uses $1024^2$ pixels at 0.439453125\,mm; intensities are clipped at the 1st/99th percentiles, mapped to 8-bit grayscale, and resized to $512^2$ without padding. Isocenter field of view is $393.75\times393.75$\,mm.

\paragraph{Detector coordinates.}
For width $n$, spacing is $\rho=450/n$ and the principal point is $(n-1)/2$. A point $(x,y,z)$ projects to
\begin{align}
u&=-\frac{1200x}{\rho(1050-y)}+u_0,\quad
v=-\frac{1200z}{\rho(1050-y)}+v_0.\label{eq:supp_projection}
\end{align}
Pixel centers map by $u_{128}=(u_{512}+0.5)/4-0.5$, and likewise for $v$. Raw AP is bilinearly resized to $128^2$; seven restored FleXray maps are area-averaged. Global and regional samplers use $(2u/127-1,2v/127-1)$ and $2[(u,v)-(u_R,v_R)]/s_R$, respectively. Both sample the same projected point with bilinear interpolation, aligned corner centers, and zero padding.

\paragraph{Isocenter and source crop.}
The NIfTI affine maps voxel centers to RAS. At approximately 4\,mm resolution, voxels above $-500$ HU define the body. Within its central lateral half and posterior 60\%, counts above 250 HU form a superior--inferior profile, smoothed by a 15\,mm Gaussian. A 400\,mm window starts at its lower end; lung/upper-lumbar intensities can shorten its upper end. Isocenter uses body midline, posterior-band center, and window midpoint shifted 8\,mm caudally. Body counts replace an unavailable bone profile. RAS crop half extents are $(276.875,180,276.875)$\,mm, clipped to scan bounds; affine translation is shifted by the isocenter.

\paragraph{Metric reference frame.}
The fixed $128\times128\times160$ grid has RAS origin $(-239,-143,-287)$\,mm and center-to-center spans $478\times286\times574$\,mm, divided by 127, 127, and 159 for spacing. Bounds come from training data. Full-resolution TotalSegmentator masks shift by the isocenter and are nearest-neighbor resampled. Classes are L1--L5, sacrum including S1 when present, and both hip bones. Scans are screened for label presence, bone volume, spacing, field of view, and metal. References are automatic masks.

\paragraph{Frozen observations.}
Frozen FleXray processes full AP and lumbar/pelvic crops without test-time augmentation. Probabilities above 0.25 define crops; the predicted sacral centroid splits the hip channel. Regions and poses use no test CT. Listed pretraining sources omit both CT sources; individual patient identifiers are unavailable.

\clearpage\onecolumn
\section{Comparison configurations and per-bone results}
\label{sec:supp_comparisons}
Every trainable stage runs 60 epochs, with seeds 0,1, and 2. An epoch traverses all eligible scan records or bone records, as appropriate to the stage. Table~\ref{tab:baseline_protocol} records the evaluated configurations; Table~\ref{tab:per_bone} retains all eight structures. Table~\ref{tab:extended_alignment} gives the complete native/rigid comparison.

\par\vspace{6 pt}
\noindent\begin{minipage}{\textwidth}
\captionsetup{type=table}\centering
\caption{Inputs, targets, training stages, and validation-selected epochs for the 15 configurations. Every trainable stage runs 60 epochs; selected checkpoints are listed in seed order.}
\label{tab:baseline_protocol}
\small\setlength{\tabcolsep}{3.5 pt}\renewcommand{\arraystretch}{1.16}
\begin{tabular}{@{}p{76 pt}p{66 pt}p{67 pt}p{165 pt}cc@{}}
\toprule Method & Input & Prediction & Configuration & Stages & Final epochs\\\midrule
Broadcast U-Net & AP & Bone labels & Broadcast 2D features into a 3D U-Net & 1 & 60/45/60 \\
Henzler & AP & CT to labels & Released coarse model without fine fusion, adapted to the RAS grid & 2 & 60/60/50 \\
PatRecon & AP & CT to labels & Author reconstruction network trained across patients & 2 & 25/20/60 \\
XctNet & AP & CT to labels & Single-AP paper reimplementation with the ResNet/New-Inception architecture & 2 & 60/60/40 \\
3DDX-Direct & AP, camera & Masks and ray depths to labels & Front/back depth and CASI, with a training depth-center term and direct voxelization & 1 & 60/60/55 \\
X2V & AP & Per-bone occupancy & ViT and conditional normalization with metric-grid queries & 1 & 50/30/35 \\
X2B & AP & Per-bone occupancy & Feed-forward ConvNeXt occupancy variant & 1 & 25/35/30 \\
SR-Net & AP & Bone labels & Single-view wavelet and dimension-transform network & 1 & 60/55/60 \\
XctDiff & AP & CT to labels & Author VQ, progressive embedding, and conditional diffusion adapted to one AP view & 4 & 60/60/60 \\
VolumeNeRF & AP, camera & CT to labels & Author generator and attention with training-only likelihood and mean CT & 2 & 60/60/15 \\
RT-SRTS & AP & Bone labels and CT & Author dual heads adapted to one AP view and nine label channels & 1 & 50/55/25 \\
Joint 2D--3D SSM & AP & Bone labels & Eight-bone training template, 2D and 3D registration, and joint PCA & 3 + PCA & Stage-specific \\
SSR/PIFu & AP, soft maps, camera & Per-point labels & Single-view pixel-aligned queries in the metric grid & 1 & 50/60/55 \\
VolumeNeRF (matched) & AP, soft maps, ROIs, camera & CT to labels & 17-channel stem and ROI encoder, adding bone CE and Dice supervision & 2 & 60/55/50 \\
Raysemble & AP, soft maps, ROIs, camera & Canonical occupancy and metric pose & Bone autoencoder, ray-conditioned shape, then frozen-shape placement & 3 & 15/35/25 \\
\bottomrule

\end{tabular}\end{minipage}

\paragraph{Supervision.}
Label/occupancy methods use CT-derived training masks. 3DDX-Direct also uses ray-depth targets. CT synthesis methods use CT targets and a training-split segmenter frozen at reconstruction time; matched VolumeNeRF adds bone cross-entropy and Dice through this segmenter. RT-SRTS predicts CT and labels. Joint SSM uses training masks for registration and PCA. XctDiff's VQ stage uses fixed VGG features for LPIPS. Raysemble, matched VolumeNeRF, and SSR/PIFu use the same frozen FleXray checkpoint. Table~\ref{tab:baseline_protocol} specifies each method's supervision and training stages.

\paragraph{What is held constant.}
Every configuration uses the same patient split, AP view, eight-label target mapping, physical evaluation grid, missing-bone rule, and patient-weighted scoring. The matched-input comparison additionally fixes the AP, soft maps, per-bone regions, and camera parameters. Table~\ref{tab:baseline_protocol} specifies the stages and sample units.

\clearpage
\noindent\begin{minipage}{\textwidth}
\captionsetup{type=table}\centering
\caption{Per-bone reconstruction before alignment. Each cell gives Dice (\%) above and HD95/ASSD (mm) below. Means average patients and three seeds; bold identifies the best value per bone.}
\label{tab:per_bone}
\fontsize{8}{9}\selectfont\setlength{\tabcolsep}{1 pt}\renewcommand{\arraystretch}{.96}
\newcommand{\distances}[2]{\makebox[20.7 pt][r]{#1}\kern.35 pt/\makebox[17 pt][r]{#2}}
\begin{tabular*}{\textwidth}{@{\extracolsep{\fill}}l*{8}{c}@{}}
\toprule Method & L1 & L2 & L3 & L4 & L5 & Sacrum & L.hip & R.hip\\\midrule
Broadcast U-Net & 46.2 & 49.3 & 52.1 & 54.1 & 52.2 & \textbf{62.4} & 67.0 & 68.2 \\
 & \distances{16.70}{6.06} & \distances{16.49}{5.54} & \distances{14.38}{5.10} & \distances{12.57}{4.70} & \distances{12.50}{4.82} & \distances{13.81}{4.90} & \distances{13.74}{4.55} & \distances{13.34}{4.42} \\
\addlinespace[1pt]
Henzler & 40.6 & 47.3 & 50.3 & 52.2 & 49.8 & 57.6 & 61.5 & 60.5 \\
 & \distances{48.29}{37.38} & \distances{16.43}{5.56} & \distances{14.95}{5.04} & \distances{13.53}{4.81} & \distances{14.47}{5.02} & \distances{16.28}{5.45} & \distances{16.28}{5.33} & \distances{18.55}{5.63} \\
\addlinespace[1pt]
PatRecon & 31.7 & 36.7 & 39.9 & 44.1 & 43.8 & 46.0 & 48.9 & 48.0 \\
 & \distances{23.04}{10.36} & \distances{29.25}{16.19} & \distances{24.18}{11.77} & \distances{28.83}{17.48} & \distances{29.02}{17.73} & \distances{28.91}{9.04} & \distances{25.46}{8.14} & \distances{29.96}{9.29} \\
\addlinespace[1pt]
XctNet & 42.4 & 47.0 & 49.2 & 52.0 & 48.9 & 58.1 & 62.0 & 61.2 \\
 & \distances{45.89}{33.56} & \distances{28.80}{17.48} & \distances{31.38}{20.83} & \distances{29.37}{20.23} & \distances{14.77}{5.44} & \distances{16.91}{5.61} & \distances{17.07}{6.04} & \distances{18.60}{5.62} \\
\addlinespace[1pt]
3DDX-Direct & 45.2 & 50.0 & 53.8 & 56.5 & 53.2 & 59.3 & 64.4 & 66.2 \\
 & \distances{20.10}{6.22} & \distances{16.85}{5.49} & \distances{16.25}{4.96} & \distances{12.44}{4.25} & \distances{13.32}{4.46} & \distances{12.45}{4.44} & \distances{11.88}{4.08} & \distances{11.46}{3.90} \\
\addlinespace[1pt]
X2V & 31.9 & 35.7 & 38.8 & 39.8 & 35.0 & 49.2 & 50.0 & 49.3 \\
 & \distances{28.99}{12.44} & \distances{27.86}{11.39} & \distances{27.04}{10.75} & \distances{22.99}{9.96} & \distances{25.82}{10.45} & \distances{23.08}{9.21} & \distances{21.52}{8.28} & \distances{22.97}{8.74} \\
\addlinespace[1pt]
X2B & 33.5 & 32.6 & 38.2 & 37.0 & 32.3 & 49.1 & 50.6 & 49.4 \\
 & \distances{28.10}{11.54} & \distances{39.96}{22.50} & \distances{25.49}{10.47} & \distances{25.00}{10.17} & \distances{27.11}{10.97} & \distances{30.93}{16.61} & \distances{21.36}{8.22} & \distances{22.68}{8.68} \\
\addlinespace[1pt]
SR-Net & 47.1 & 50.3 & 51.5 & 51.6 & 46.7 & 60.5 & 62.1 & 64.7 \\
 & \distances{19.24}{6.39} & \distances{17.43}{5.74} & \distances{16.54}{5.50} & \distances{14.38}{5.27} & \distances{14.39}{5.65} & \distances{13.72}{4.91} & \distances{15.03}{4.84} & \distances{13.94}{4.73} \\
\addlinespace[1pt]
XctDiff & 18.0 & 25.0 & 29.2 & 27.2 & 22.2 & 34.0 & 48.2 & 45.9 \\
 & \distances{66.55}{49.20} & \distances{45.98}{28.81} & \distances{24.42}{8.79} & \distances{24.55}{8.62} & \distances{31.99}{14.61} & \distances{24.80}{7.77} & \distances{23.59}{7.97} & \distances{48.35}{10.46} \\
\addlinespace[1pt]
VolumeNeRF & 37.2 & 41.5 & 43.6 & 44.7 & 43.3 & 50.6 & 52.7 & 52.1 \\
 & \distances{27.53}{10.09} & \distances{26.41}{8.89} & \distances{22.42}{8.48} & \distances{19.20}{7.78} & \distances{19.69}{7.83} & \distances{20.98}{7.37} & \distances{25.54}{7.75} & \distances{25.14}{8.03} \\
\addlinespace[1pt]
RT-SRTS & 13.7 & 17.0 & 26.9 & 23.8 & 16.6 & 41.0 & 51.5 & 51.3 \\
 & \distances{234.63}{75.55} & \distances{178.00}{49.18} & \distances{46.46}{25.06} & \distances{110.38}{16.25} & \distances{225.31}{43.88} & \distances{152.53}{22.00} & \distances{75.53}{14.79} & \distances{77.65}{10.94} \\
\addlinespace[1pt]
Joint 2D--3D SSM & 29.1 & 33.1 & 34.6 & 35.4 & 34.8 & 36.0 & 40.8 & 44.1 \\
 & \distances{35.87}{9.53} & \distances{30.85}{8.30} & \distances{27.65}{7.46} & \distances{25.46}{6.92} & \distances{25.33}{6.92} & \distances{30.09}{7.83} & \distances{27.80}{7.76} & \distances{24.11}{7.25} \\
\addlinespace[1pt]
SSR/PIFu & 44.4 & 47.3 & 49.7 & 49.3 & 50.7 & 60.4 & 67.4 & 66.7 \\
 & \distances{17.88}{6.87} & \distances{17.31}{6.17} & \distances{16.71}{5.92} & \distances{18.48}{6.09} & \distances{16.41}{5.84} & \distances{19.93}{6.34} & \distances{13.99}{4.64} & \distances{14.02}{4.86} \\
\addlinespace[1pt]
VolumeNeRF (matched) & 54.5 & 57.4 & 59.7 & 62.0 & 57.6 & 59.8 & 62.0 & 62.1 \\
 & \distances{28.55}{7.46} & \distances{21.52}{7.10} & \distances{18.37}{5.42} & \distances{13.27}{4.01} & \distances{16.04}{4.74} & \distances{19.45}{6.06} & \distances{20.45}{5.70} & \distances{19.43}{5.63} \\
\addlinespace[1pt]
\textbf{Raysemble} & \textbf{61.6} & \textbf{66.1} & \textbf{67.4} & \textbf{66.4} & \textbf{59.3} & 61.9 & \textbf{69.2} & \textbf{70.1} \\
 & \distances{\textbf{9.18}}{\textbf{3.34}} & \distances{\textbf{8.02}}{\textbf{2.82}} & \distances{\textbf{8.37}}{\textbf{2.73}} & \distances{\textbf{8.17}}{\textbf{2.86}} & \distances{\textbf{9.32}}{\textbf{3.61}} & \distances{\textbf{10.88}}{\textbf{4.18}} & \distances{\textbf{9.28}}{\textbf{3.45}} & \distances{\textbf{9.46}}{\textbf{3.59}} \\
\addlinespace[1pt]
\bottomrule

\end{tabular*}\end{minipage}
\par\vspace{10 pt}
\noindent\begin{minipage}{\textwidth}
\captionsetup{type=table}\centering
\caption{Complete native and rigid results on 43 CT scans from 23 patients. Means $\pm$ sample standard deviations across three seeds. Rigid alignment removes whole-case rotation/translation only. Missing predictions count bones and affected outputs over 1,032 reference instances and 129 scan--seed outputs per method.}
\label{tab:extended_alignment}\label{tab:missing_bones}
\fontsize{8}{9}\selectfont\setlength{\tabcolsep}{3 pt}\renewcommand{\arraystretch}{1.12}
\begin{tabular*}{\textwidth}{@{\extracolsep{\fill}}lrrrrrrcc@{}}
\toprule &\multicolumn{3}{c}{Native: no alignment}&\multicolumn{3}{c}{Rigid: no scaling}&\multicolumn{2}{c}{Missing predictions}\\
\cmidrule(lr){2-4}\cmidrule(lr){5-7}\cmidrule(l){8-9}
Method & Dice (\%) & HD95 (mm)& ASSD (mm)& Dice (\%)&HD95 (mm)&ASSD (mm)&Bones&Outputs\\\midrule
Broadcast U-Net & $56.45\pm1.17$ & $14.19\pm2.40$ & $5.01\pm0.63$ & $66.22\pm2.88$ & $11.39\pm2.39$ & $3.78\pm0.65$ & 0 & 0 \\
Henzler & $52.46\pm0.53$ & $19.85\pm4.19$ & $9.28\pm4.27$ & $59.42\pm0.57$ & $17.04\pm3.88$ & $8.24\pm4.22$ & 6 & 6 \\
PatRecon & $42.39\pm6.25$ & $27.33\pm8.77$ & $12.50\pm5.44$ & $49.73\pm6.32$ & $22.64\pm7.50$ & $10.33\pm4.80$ & 9 & 6 \\
XctNet & $52.60\pm2.16$ & $25.35\pm12.62$ & $14.35\pm12.46$ & $59.29\pm2.21$ & $22.59\pm12.47$ & $13.30\pm12.54$ & 16 & 7 \\
3DDX-Direct & $56.08\pm0.95$ & $14.35\pm0.95$ & $4.73\pm0.11$ & $60.85\pm0.52$ & $12.10\pm1.04$ & $4.01\pm0.21$ & 0 & 0 \\
X2V & $41.21\pm1.23$ & $25.03\pm0.80$ & $10.15\pm0.20$ & $55.66\pm1.13$ & $17.05\pm1.19$ & $5.85\pm0.14$ & 0 & 0 \\
X2B & $40.35\pm1.19$ & $27.58\pm4.39$ & $12.39\pm4.03$ & $53.78\pm2.64$ & $19.62\pm5.15$ & $8.49\pm4.42$ & 5 & 3 \\
SR-Net & $54.32\pm1.30$ & $15.58\pm1.28$ & $5.38\pm0.27$ & $65.51\pm2.11$ & $11.76\pm0.87$ & $3.66\pm0.23$ & 0 & 0 \\
XctDiff & $31.22\pm8.82$ & $36.28\pm16.74$ & $17.03\pm10.62$ & $37.63\pm12.65$ & $31.81\pm17.50$ & $15.06\pm11.10$ & 12 & 10 \\
VolumeNeRF & $45.72\pm5.65$ & $23.36\pm2.13$ & $8.28\pm1.02$ & $55.46\pm2.40$ & $17.52\pm3.26$ & $5.69\pm0.77$ & 0 & 0 \\
RT-SRTS & $30.23\pm14.20$ & $137.56\pm63.21$ & $32.21\pm16.28$ & $21.68\pm22.55$ & $104.69\pm69.69$ & $31.78\pm21.61$ & 4 & 4 \\
Joint 2D--3D SSM & $35.99\pm1.99$ & $28.40\pm4.13$ & $7.75\pm0.80$ & $44.11\pm3.59$ & $22.37\pm2.67$ & $5.82\pm0.65$ & 0 & 0 \\
SSR/PIFu & $54.48\pm0.21$ & $16.84\pm0.21$ & $5.84\pm0.03$ & $56.31\pm0.17$ & $14.82\pm0.23$ & $5.57\pm0.03$ & 0 & 0 \\
VolumeNeRF (matched) & $59.38\pm0.30$ & $19.63\pm6.49$ & $5.76\pm1.25$ & $62.88\pm1.64$ & $16.44\pm5.36$ & $5.02\pm1.14$ & 0 & 0 \\
\textbf{Raysemble} & $\mathbf{65.24\pm0.46}$ & $\mathbf{9.09\pm0.15}$ & $\mathbf{3.32\pm0.06}$ & $\mathbf{73.58\pm0.09}$ & $\mathbf{6.96\pm0.03}$ & $\mathbf{2.49\pm0.01}$ & 0 & 0 \\
\bottomrule

\end{tabular*}\end{minipage}

\clearpage
\section{Placement with local shape held fixed}
\label{sec:supp_placement}
Initial, residual, and absolute variants share each seed's decoded shapes. Residual placement selects epoch 0; absolute placement selects epochs 15, 35, and 25. Depth error is measured along RAS $y$.
\par\vspace{0pt}
\noindent\begin{minipage}{\textwidth}\captionsetup{type=figure}\centering
\includegraphics[width=.90\textwidth]{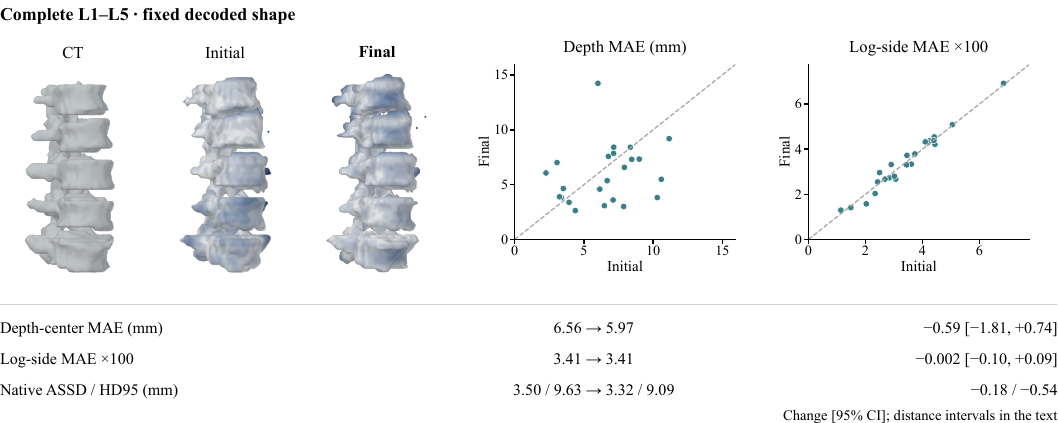}
\caption{Fixed-shape placement. Complete L1--L5 shares one physical window. Scatterplots show 23 patient means; below the diagonal indicates improvement. Depth and scale changes have 95\% paired bootstrap intervals. Distance-change intervals are $[-0.42,+0.07]$\,mm (ASSD) and $[-1.19,+0.14]$\,mm (HD95).}
\label{fig:placement}\end{minipage}
\par\vspace{0pt}
\section{Patient-level gains with matched inputs}
\label{sec:supp_patient_gains}
Across 23 paired patients, native Dice improves for 18 and both surface distances improve for all 23. Mean gains are 5.86 percentage points, 10.55\,mm HD95, and 2.44\,mm ASSD.
\noindent\begin{minipage}{\textwidth}\captionsetup{type=figure}\centering
\includegraphics[width=.93\textwidth]{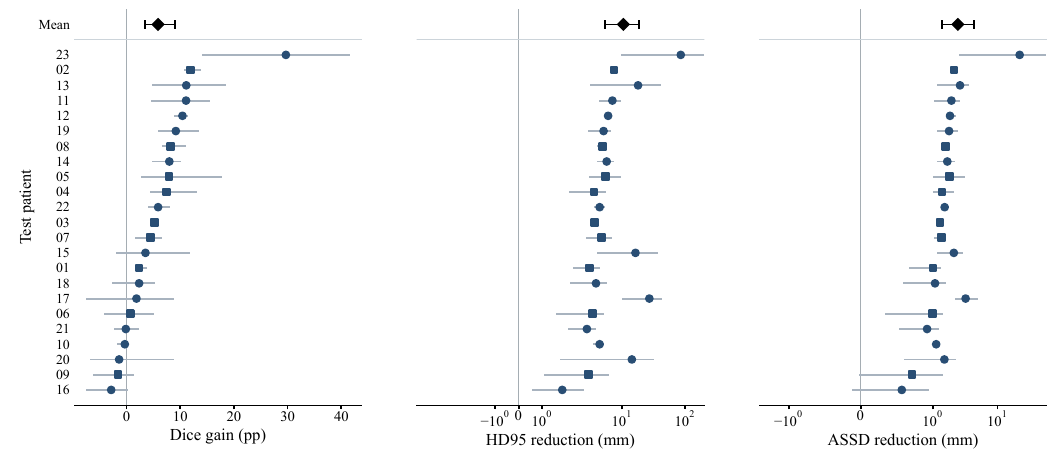}
\caption{Paired gains for 23 patients after averaging scans and seeds. Thin segments show the three-seed range; black diamonds and bars show the patient mean and 95\% bootstrap interval (10,000 resamples). Circles/squares denote CT-TRI/AbdomenCT-1K. Positive values favor Raysemble; distance axes use symmetric-log scaling.}
\label{fig:patient_pairs}\end{minipage}
\par\vspace{0pt}
\section{Class presence and CT-segmentation diagnostic}
\label{sec:supp_failures}
Table~\ref{tab:missing_bones} counts absent classes; matched VolumeNeRF and Raysemble retain all eight classes in all 129 outputs. Reference CT through matched VolumeNeRF's frozen segmenter yields $91.82\pm0.31$\% Dice, $4.59\pm0.20$\,mm HD95, and $0.76\pm0.03$\,mm ASSD. This comparison characterizes the segmenter on acquired CT.
\clearpage
\onecolumn
\section{Clinical transfer on paired pelvic radiographs}
\label{sec:supp_ethics}
The two retrospective cases were collected at the First People's Hospital of Dali City, with 16- and 70-day X-ray--CT intervals. The study was approved by the hospital's Institutional Review Board (approval ID: SYYLL-2024001), which waived informed consent because of the retrospective design. The 16-day case was used during development.
\par\vspace{5 pt}
\noindent\begin{minipage}{\textwidth}
\centering
\captionsetup{type=figure}
\includegraphics[width=\textwidth]{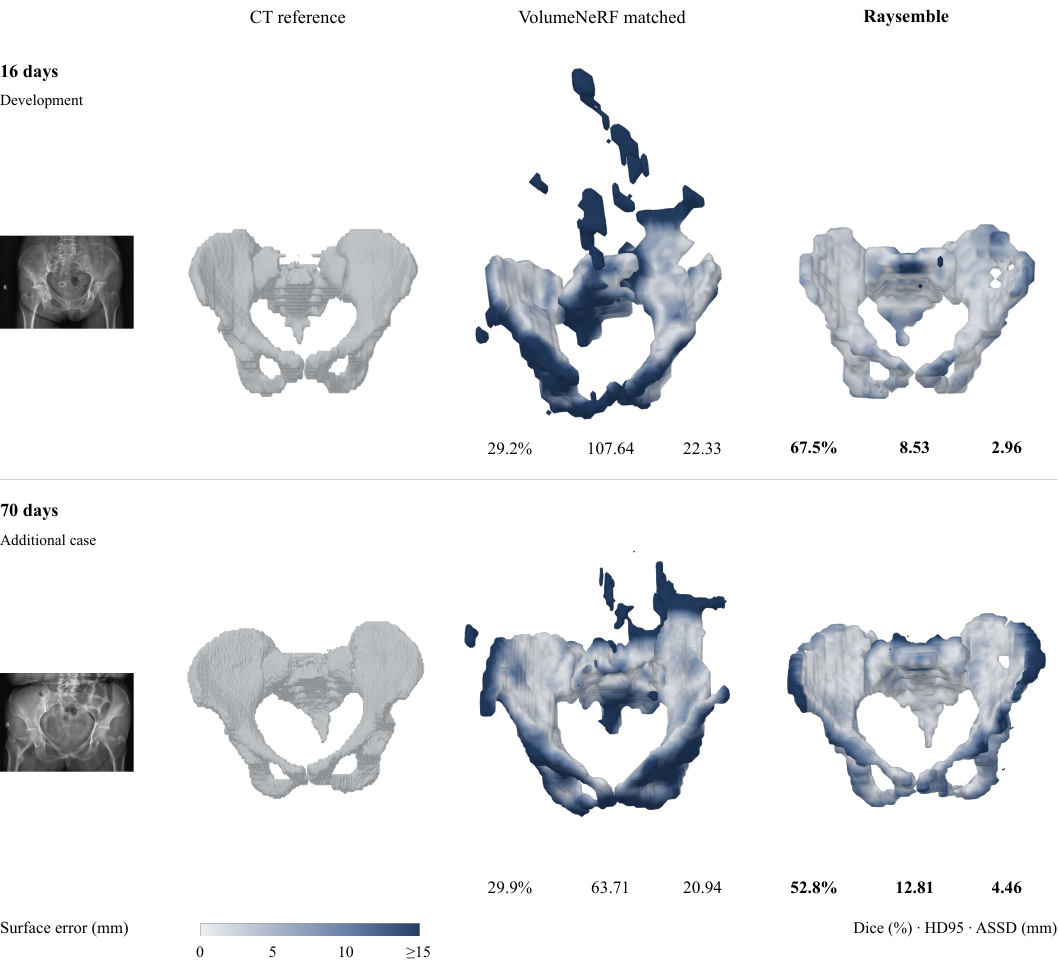}
\caption{Paired clinical pelvic reconstructions. Each case uses a shared physical frame for CT, matched-input VolumeNeRF, and Raysemble. Numbers are three-seed means of pelvic Dice (\%), HD95 (mm), and ASSD (mm) after one whole-pelvis rigid alignment without scaling; best values are bold.}
\label{fig:clinical_extended}
\end{minipage}
\clearpage
\onecolumn
\section{Whole anatomy in native and aligned coordinates}

\noindent\begin{minipage}{\textwidth}
\centering
\captionsetup{type=figure}
\includegraphics[width=.94\textwidth]{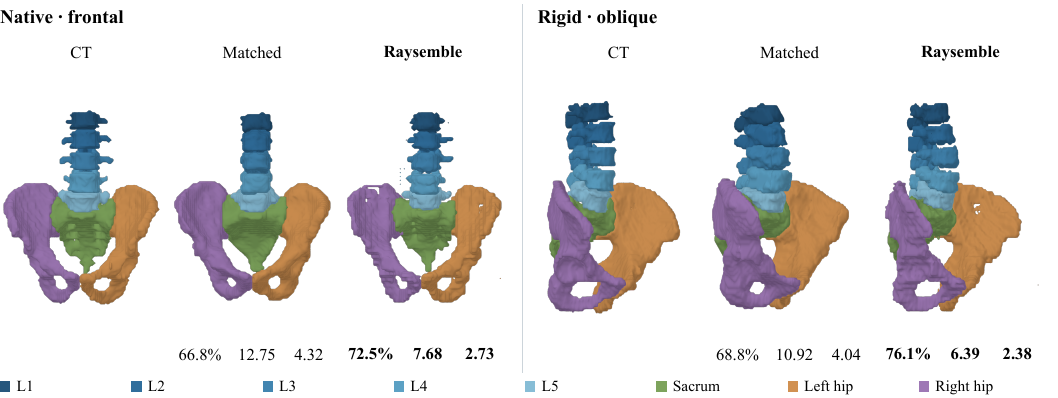}
\caption{A single CT-TRI case in native frontal and rigidly aligned oblique views. CT, matched-input VolumeNeRF, and Raysemble share a physical frame within each triplet. Colors identify the eight bones. Surfaces use seed 0; Dice (\%), HD95, and ASSD (mm) average three seeds.}
\label{fig:whole_seeds}
\end{minipage}
\par\vspace{0pt}
\section{Complete local surfaces and CT correspondence}

\noindent\begin{minipage}{\textwidth}
\centering
\captionsetup{type=figure}
\includegraphics[width=.96\textwidth]{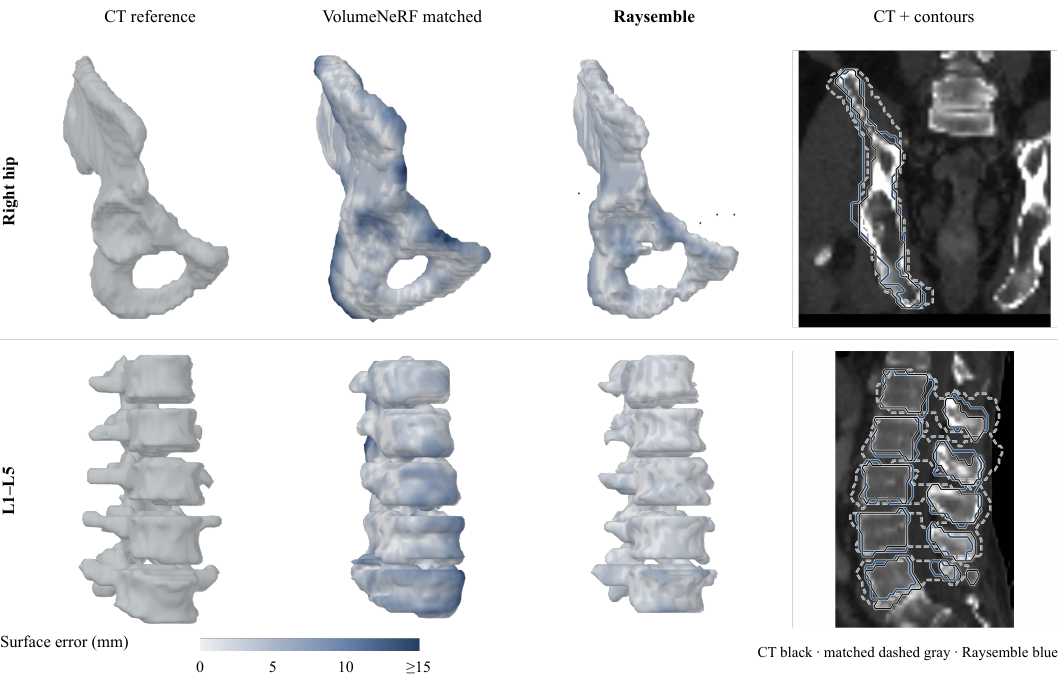}
\caption{Right hip and L1--L5 from one AbdomenCT-1K case. Surfaces share a physical window after whole-case rigid alignment. Coronal and sagittal CT sections use the same planes and crop bounds for every contour: CT in black, matched-input VolumeNeRF in dashed gray, and Raysemble in blue. Surface-error maps use 0--15\,mm.}
\label{fig:local_complete}
\end{minipage}
\clearpage
\onecolumn
\section{Depth profiles and surface-distance distributions}

\noindent\begin{minipage}{\textwidth}
\centering
\captionsetup{type=figure}
\includegraphics[width=\textwidth]{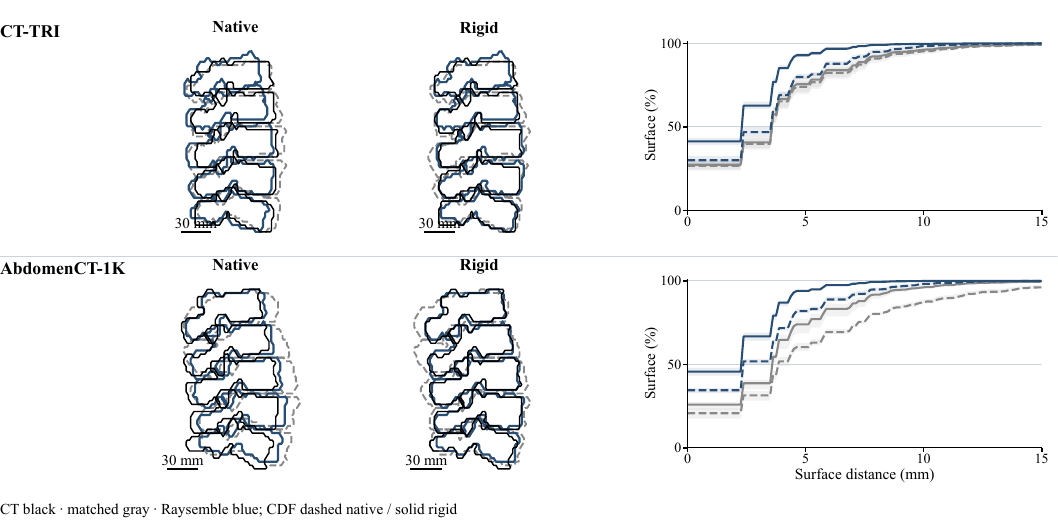}
\caption{Lumbar depth profiles and symmetric surface-distance distributions for both CT sources. CT is black, matched-input VolumeNeRF gray, and Raysemble blue. Dashed/solid CDF curves denote native/rigid evaluation, averaging eight bones, both directions, and three seeds. Bands show seed ranges. CDFs include all surface points and display thresholds up to 15\,mm.}
\label{fig:depth_distance}
\end{minipage}
\par\vspace{0pt}
\section{Training convergence and ablation behavior}

\noindent\begin{minipage}{\textwidth}
\centering
\captionsetup{type=figure}
\includegraphics[width=\textwidth]{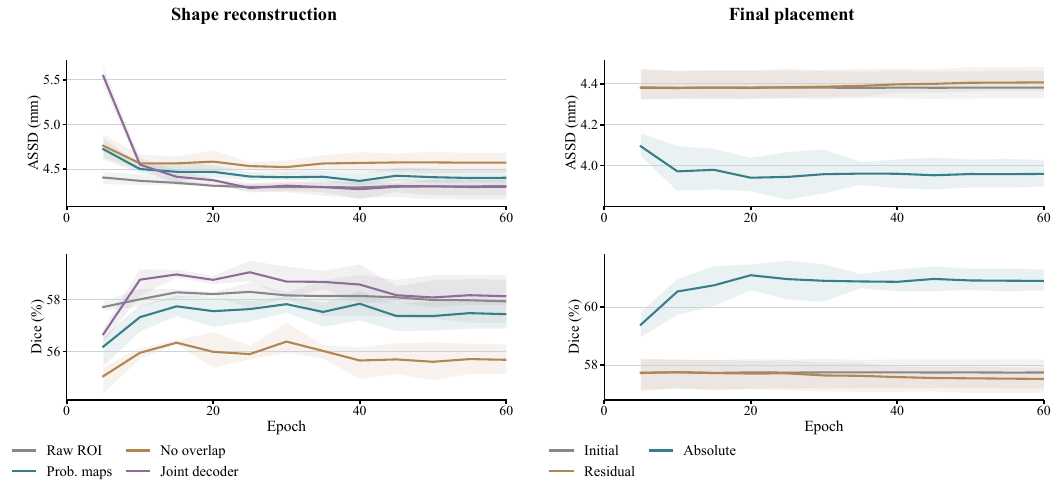}
\caption{Validation convergence over 60 epochs per stage. Lines show three-seed means and bands their ranges. The joint-decoder variant also changes mean-latent initialization and latent supervision. Checkpoint selection minimizes validation ASSD; residual placement selects epoch 0. Test-set ablations are reported in the main paper.}
\label{fig:convergence}
\end{minipage}
\clearpage\onecolumn
\section{Anatomical probabilities and regional observations}
\label{sec:probability_evidence}
\noindent\begin{minipage}{\textwidth}\captionsetup{type=figure}\centering
\includegraphics[width=\textwidth]{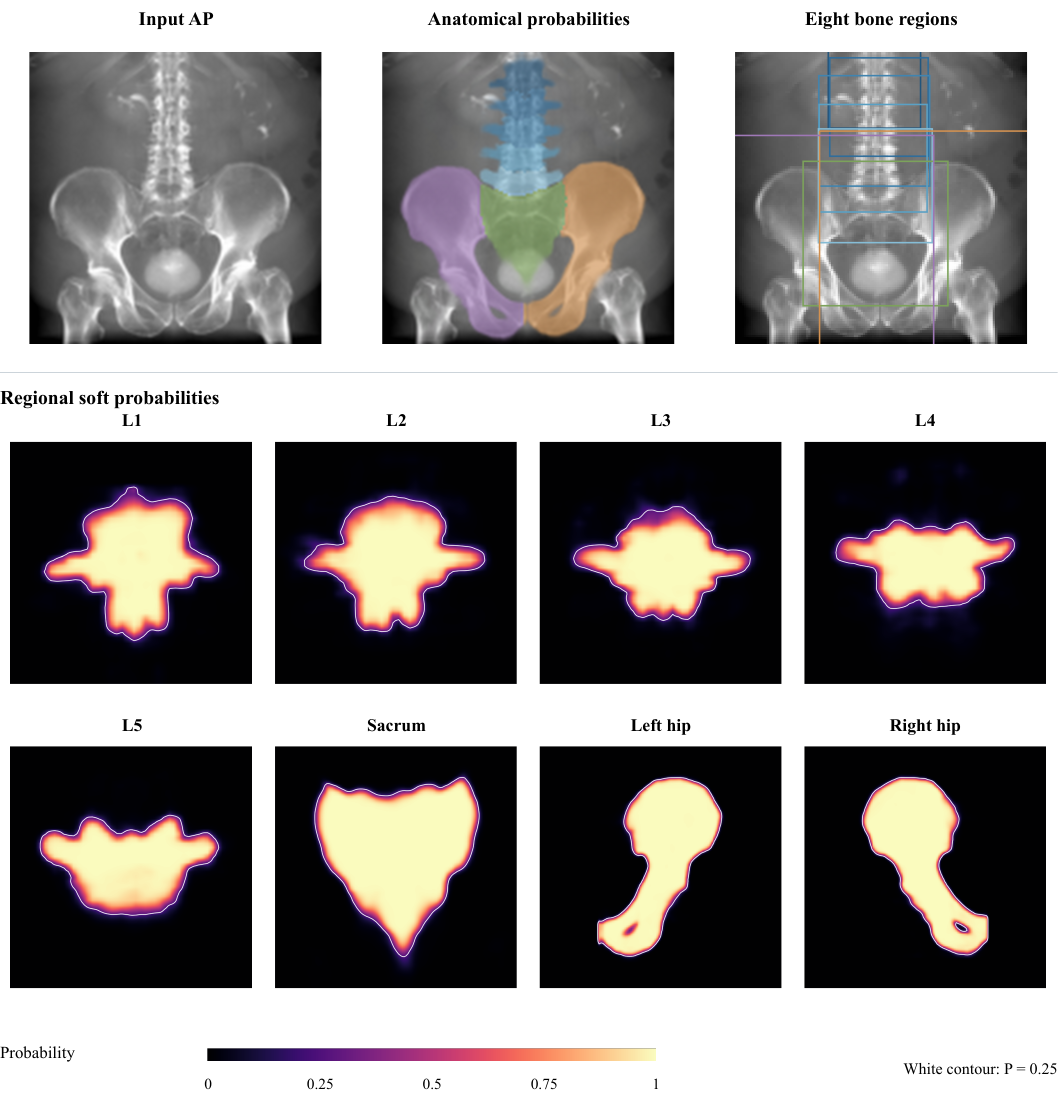}
\caption{Detector-space probabilities and per-bone ROIs for one CT-TRI case. The seven FleXray maps yield eight observations after splitting the hip channel at the predicted sacral centroid. Top: AP input, an anatomy overlay using the largest probability at each pixel, and the square model ROIs. Bottom: the actual regional soft maps on a fixed 0--1 color scale. White contours mark the 0.25 threshold used to define regions.}
\label{fig:probability_evidence}\end{minipage}
\clearpage\onecolumn
\section{From regional evidence to bone surfaces}
\noindent\begin{minipage}{\textwidth}\captionsetup{type=figure}\centering
\includegraphics[width=\textwidth]{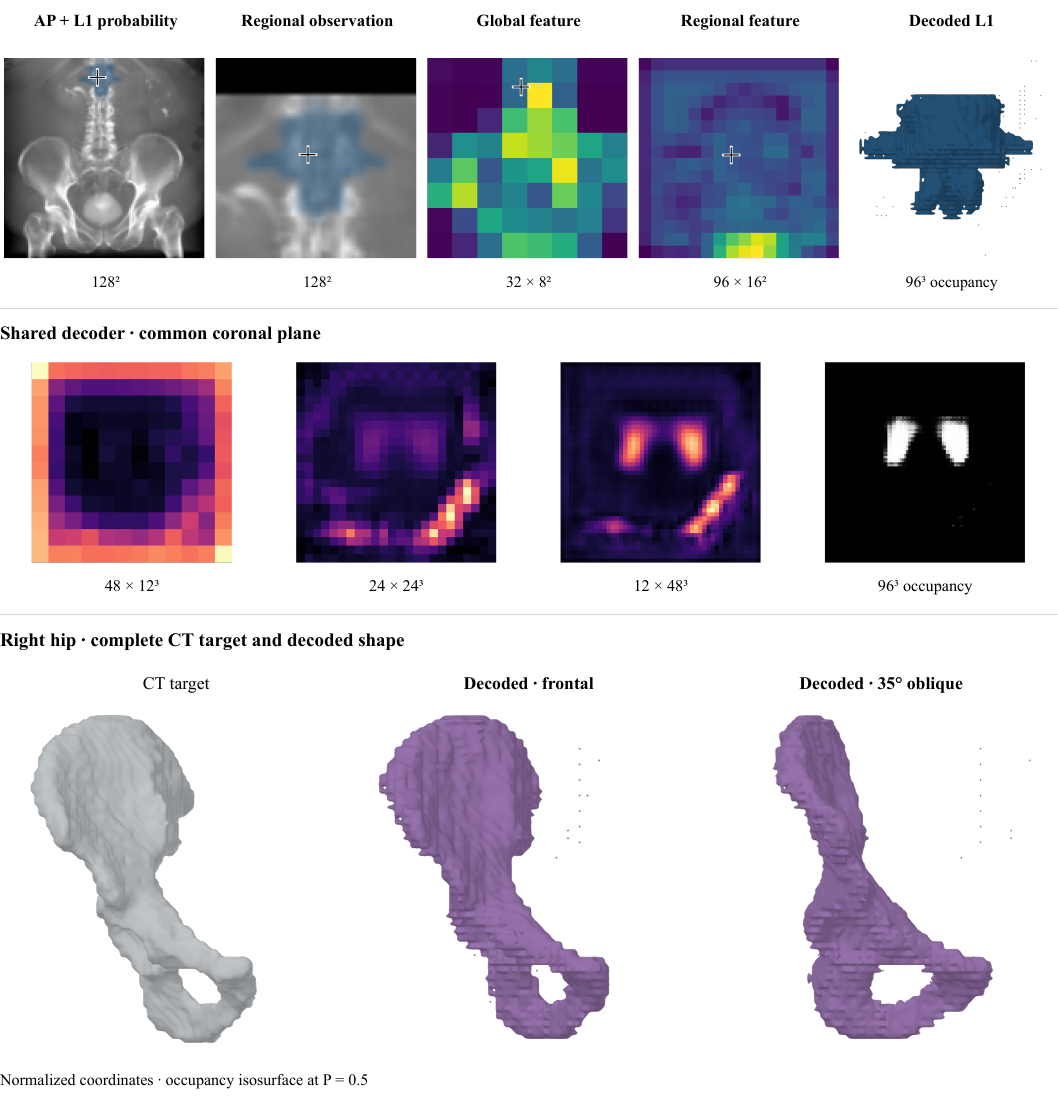}
\caption{Forward-pass activations for L1 and the right hip from the seed-0 checkpoint on the same CT-TRI case as Fig.~\ref{fig:probability_evidence}. Crosses mark one projected query in global and ROI coordinates. AP overlays locate the query; the encoders process soft probability maps. Feature colors encode normalized channel RMS. Decoder responses progress through $12^3$, $24^3$, and $48^3$ to $96^3$ occupancy on a common coronal plane. The lower row compares the normalized CT hip target with anterior and oblique views of the decoded isosurface at probability 0.5.}
\label{fig:network_evidence}\end{minipage}
\clearpage\onecolumn
\section{Multiple-method reconstruction for individual bones}\label{sec:multi_bone_visual}
\noindent\begin{minipage}{\textwidth}\captionsetup{type=figure}\centering
\includegraphics[width=.98\textwidth]{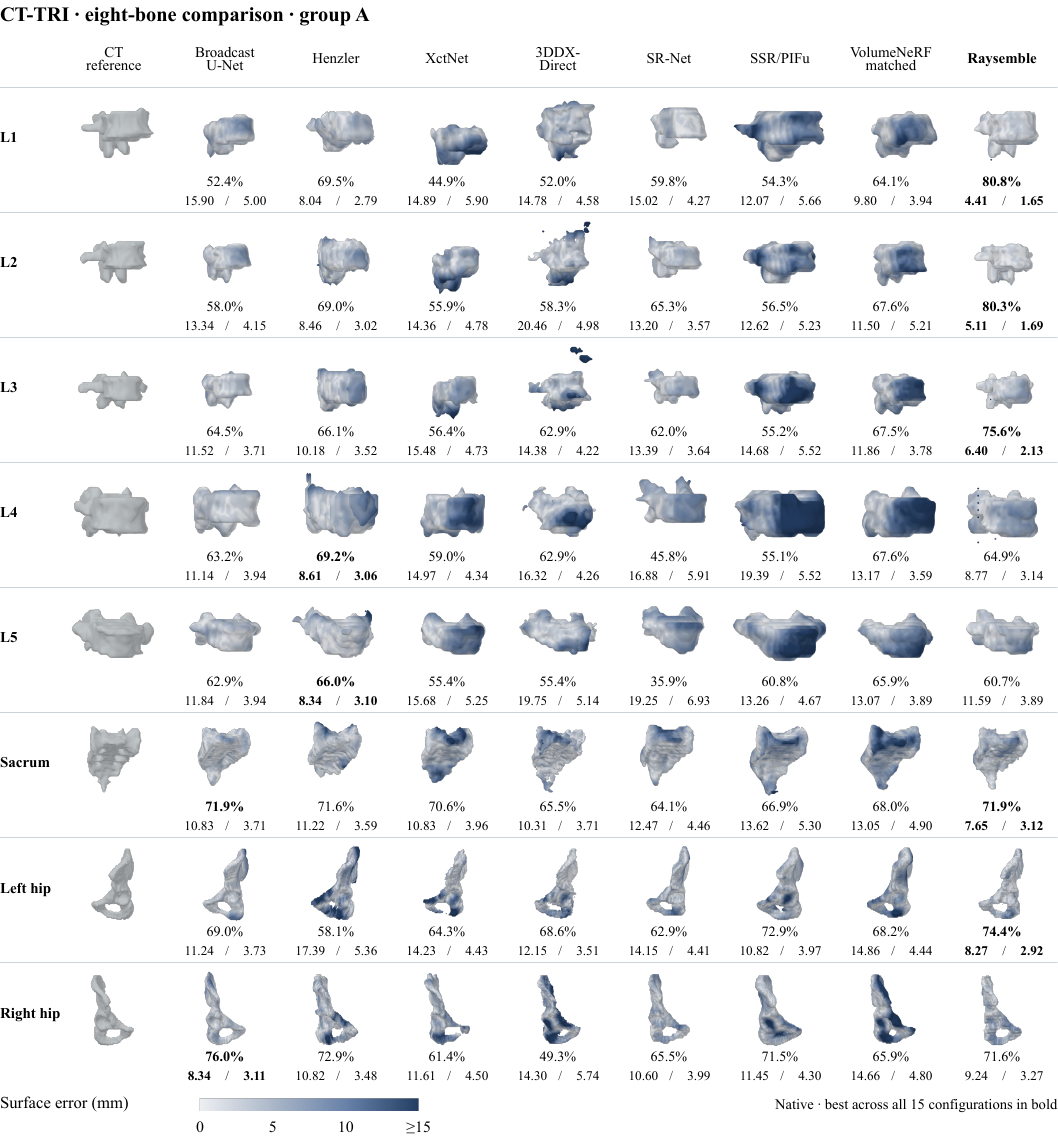}
\caption{Eight-bone reconstruction on CT-TRI, group A. This page and the following group-B page cover all 14 comparators and Raysemble, with CT and Raysemble repeated for reference. Unmarked columns share a physical frame per bone row; dagger-marked columns use expanded viewports where needed. Surfaces show seed 0; native Dice (\%) and HD95/ASSD (mm) average three seeds. Bold marks the best rounded mean across all 15 configurations. Surface errors use 0--15\,mm.}
\label{fig:bone_atlas_cttri_A}\label{fig:bone_atlas_cttri}\end{minipage}

\clearpage\onecolumn
\section*{Individual-bone comparison: CT-TRI (group B)}
\noindent\begin{minipage}{\textwidth}\captionsetup{type=figure}\centering
\includegraphics[width=.98\textwidth]{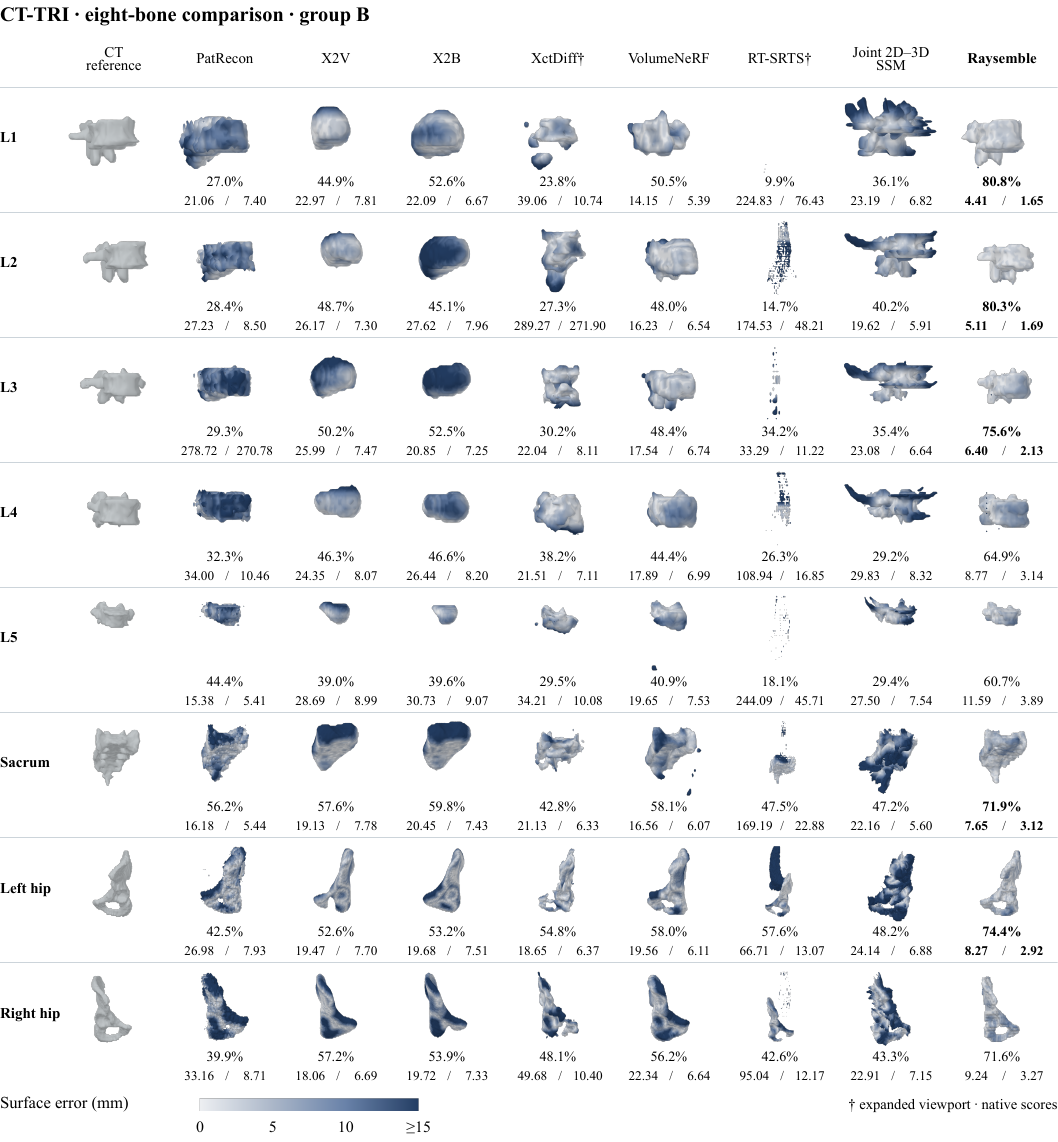}
\caption{Eight-bone reconstruction on CT-TRI, group B, completing the methods in Fig.~\ref{fig:bone_atlas_cttri}. Dagger-marked columns use expanded viewports to show displaced components; unmarked columns share the row frame. Native three-seed scores and best-value formatting follow the companion page.}
\label{fig:bone_atlas_cttri_B}\end{minipage}

\clearpage\onecolumn
\section*{Individual-bone comparison: AbdomenCT-1K (group A)}
\noindent\begin{minipage}{\textwidth}\captionsetup{type=figure}\centering
\includegraphics[width=.98\textwidth]{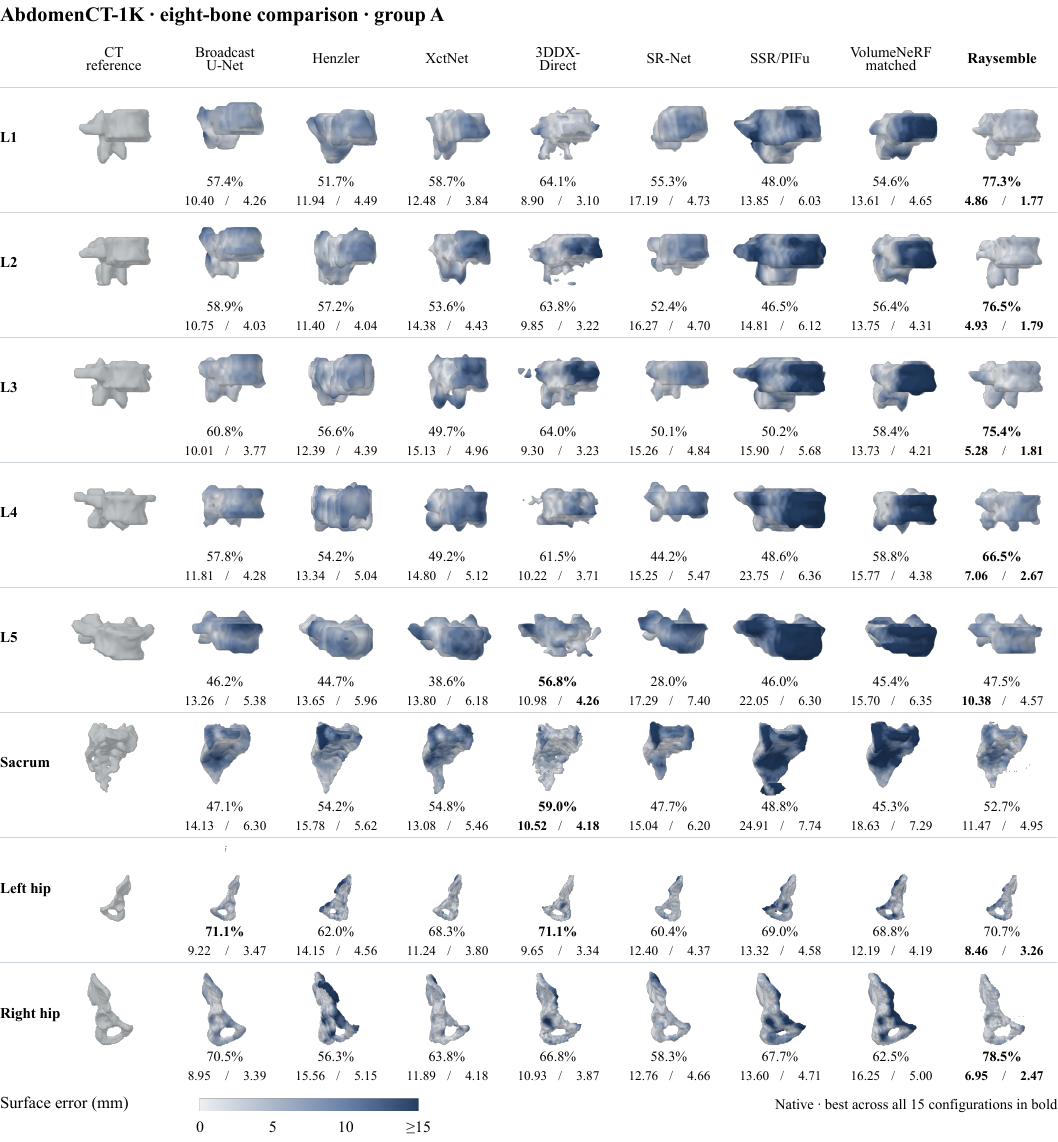}
\caption{Eight-bone reconstruction on AbdomenCT-1K, group A. This page and the following group-B page cover all 14 comparators and Raysemble, with CT and Raysemble repeated for reference. Unmarked columns share a physical frame per bone row; dagger-marked columns use expanded viewports where needed. Surfaces show seed 0; native Dice (\%) and HD95/ASSD (mm) average three seeds. Bold marks the best rounded mean across all 15 configurations. Surface errors use 0--15\,mm.}
\label{fig:bone_atlas_abdomen_A}\label{fig:bone_atlas_abdomen}\end{minipage}

\clearpage\onecolumn
\section*{Individual-bone comparison: AbdomenCT-1K (group B)}
\noindent\begin{minipage}{\textwidth}\captionsetup{type=figure}\centering
\includegraphics[width=.98\textwidth]{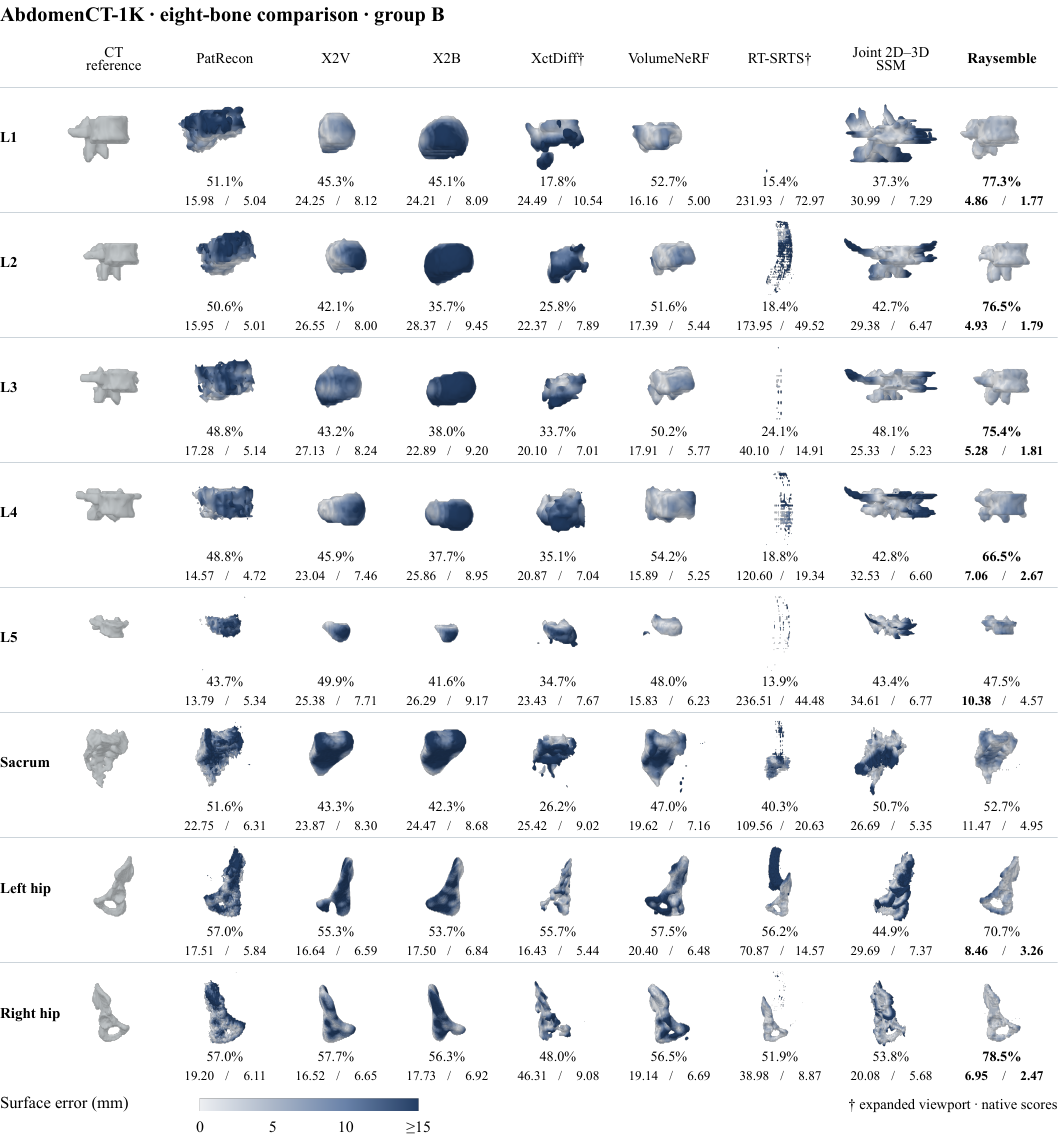}
\caption{Eight-bone reconstruction on AbdomenCT-1K, group B, completing the methods in Fig.~\ref{fig:bone_atlas_abdomen}. Dagger-marked columns use expanded viewports to show displaced components; unmarked columns share the row frame. Native three-seed scores and best-value formatting follow the companion page.}
\label{fig:bone_atlas_abdomen_B}\end{minipage}
\clearpage
\section{Per-bone localization effects and patient variation}
\label{sec:bone_population}
\noindent\begin{minipage}{\textwidth}\captionsetup{type=figure}\centering
\includegraphics[width=.98\textwidth]{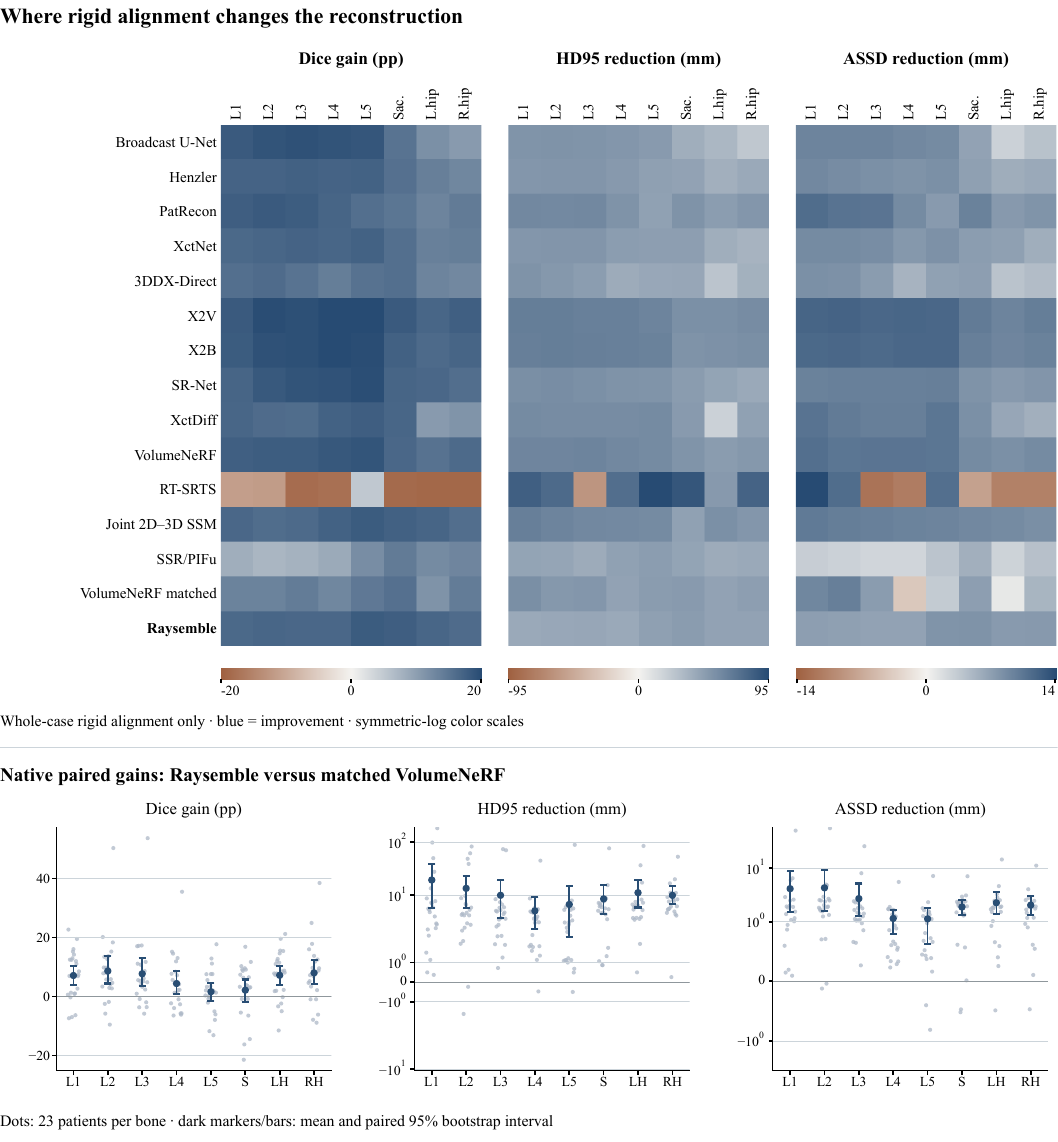}
\caption{Top: per-bone changes after one whole-case rigid transform for all 15 configurations. Positive Dice gains and distance reductions indicate improvement; colors use symmetric-log scales. Alignment applies rotation and translation. Bottom: native paired gains over matched-input VolumeNeRF. Light dots are 23 patient means per bone; dark markers and bars show means and 95\% paired bootstrap intervals (10,000 resamples). Scores average scans within patients and three seeds.}
\label{fig:bone_population}\end{minipage}
\clearpage\twocolumn
\section{Network dimensions and reproducible training}
\label{sec:implementation}
\paragraph{Tensor flow.}
The global CNN maps seven $128^2$ soft maps to $32\times 8^2$ features and a 64-dimensional summary. The regional CNN takes the seven maps plus a current-bone observation cropped to $128^2$, producing $96\times 16^2$ features. A normalized $12^3$ query lattice is placed by the initial center and cube side; one AP projection supplies both regional-relative and global sampling coordinates.

At each query, 96 regional and 32 global channels concatenate with a 16-dimensional bone embedding, three normalized coordinates, and eight mean-latent channels: 155 channels total. The residual shape network maps 155 to 48 to 32 to 8 channels. Its output adds to the mean latent. An eight-channel decoder identity gives 16 decoder input channels; progressive decoding traverses $12^3$, $24^3$, $48^3$, and $96^3$, followed by sigmoid.

\paragraph{Placement and assembly.}
The final placement stage takes the AP plus seven probability maps. Joint and bone-specific heads update detector center, depth, and log side. Back-projection converts the detector center and depth to RAS. Magnification correction scales the initial side by $(1050-d_1)/(1050-d_0)$, together with the bounded log-scale residual. The decoder's normalized occupancy is held fixed during this stage. Local occupancies and final metric poses are separate assembly inputs. Trilinear resampling uses the same RAS grid for every bone. We clip cube sides to 20--450\,mm; the highest occupancy determines the label only when it exceeds 0.5.

\paragraph{Original staged optimization.}
Each seed trains the CT shape autoencoder, then the ray-conditioned shape predictor, then final placement for 60 epochs each. The autoencoder and shape stages use 2-example microbatches accumulated to an effective batch of 4 bone records. AdamW uses initial learning rates of $2\times 10^{-4}$ for the autoencoder and $10^{-4}$ for each subsequent stage. Weight decay is $10^{-4}$ for shape training and $10^{-3}$ for final placement. A five-epoch linear warmup is followed by cosine decay to 1\% of the initial rate. Checkpoints follow the validation rules in Sec.~S2. There are 2,669 eligible training bone records, or 160,140 record presentations per complete 60-epoch bone stage. Each epoch traverses all eligible records once.

\newpage
\paragraph{Reference and prediction roles.}
Training CT labels supervise canonical occupancies and physical poses. At inference, the model uses the AP image, frozen probability maps, derived regions, bone identities, and camera constants. Test CT labels define evaluation targets and comparison windows. The reference-CT diagnostic in Sec.~S5 evaluates the segmenter. Patient splits and checkpoints follow the validation protocol.

\paragraph{Selection and aggregation.}
Checkpoints are selected on validation data. Every five epochs, end-to-end models are scored by native ASSD, first averaged within patient and then equally across patients. The shape autoencoder uses canonical validation Dice; SSM stages use their own validation objectives. Test scores average bones, scans within patient, and patients, in that order. Standard deviations describe the three seed means. Individual qualitative surfaces use seed 0; the numbers below them are the corresponding three-seed case means unless explicitly labeled otherwise.

\paragraph{Alignment and distance conventions.}
Native scores use the predicted RAS placement without fitting to CT. Rigid scores use one class-aware whole-skeleton rotation and translation, with no scaling or independent bone registration. Surface distances use physical coordinates on the fixed label grid. ASSD averages the two directional mean nearest-surface distances; HD95 is the larger of their directional 95th percentiles. A missing predicted reference bone receives Dice 0 and the physical grid diagonal for each distance. Visual heatmaps show prediction-to-reference, same-bone nearest-vertex distances; quantitative scores use symmetric voxel surfaces.

\paragraph{Qualitative rendering.}
Metric-space surface displays use common rendering settings and smoothing bounded at 0.65\,mm. Error colors encode same-bone nearest-vertex distances over 0--15\,mm; reported metrics use the original label volumes and symmetric voxel surfaces. Unmarked atlas columns share physical windows, and dagger-marked columns use expanded viewports. Canonical decoder shapes are shown in normalized coordinates with a common camera scale.

\end{document}